\documentclass[a4paper,fleqn]{cas-dc}

\usepackage[numbers,sort&compress]{natbib}

\usepackage[utf8]{inputenc}
\usepackage{amsthm}
\usepackage{import}
\usepackage{algorithm,algorithmic}
\usepackage{lipsum}
\usepackage{color} 
\usepackage{multirow}
\usepackage{graphicx}
\usepackage{float}
\usepackage{subfigure}
\usepackage{booktabs}
\usepackage{multicol}
\usepackage{multirow}
\usepackage[normalem]{ulem}
\usepackage{adjustbox}
\usepackage{emptypage}
\usepackage{amsfonts}
\usepackage{amssymb}
\usepackage[figuresright]{rotating}
\usepackage{hhline}
\usepackage{diagbox}
\usepackage{booktabs}
\usepackage{makecell}
\usepackage{subfigure} 

\usepackage[normalem]{ulem}
\useunder{\uline}{\ul}{}
\usepackage{caption}
\theoremstyle{plain}

\theoremstyle{plain}

\def\tsc#1{\csdef{#1}{\textsc{\lowercase{#1}}\xspace}}
\tsc{WGM}
\tsc{QE}

\begin{document}
\let\WriteBookmarks\relax
\def\floatpagepagefraction{1}
\def\textpagefraction{.001}
\let\printorcid\relax
%\shorttitle{<short title of the paper for running head>}    

%\shortauthors{<short author list for running head>}  

  \title[mode = title]{Semi-supervised multi-view concept decomposition}

%\tnotemark[1,2]
%
%\tnotetext[1]{This document is the results of the research project
%	funded by the National Science Foundation.}
%
%\tnotetext[2]{The second title footnote which is a longer text
%	matter to fill through the whole text width and overflow into
%	another line in the footnotes area of the first page.}
\author[]{Qi Jiang}
\ead{qi.jiang.gdut@qq.com} 

%\author[1]{Zhenhao Huang}
%\ead{zhhuang.gdut@qq.com} 

\author[]{Guoxu Zhou}
\cormark[1]
\ead{gx.zhou@gdut.edu.cn} 

\author[]{Qibin Zhao}
\cormark[2]
\ead{qibin.zhao@riken.jp} 

\address[]{School of Automation, Guangdong University of Technology, Guangzhou 510006, China}
%\address[2]{Key Laboratory of Intelligent Detection and The Internet of Things in Manufacturing, Ministry of Education, Guangzhou 510006, China}
\cortext[cor1, cor2]{Corresponding author}

\begin{abstract}[S U M M A R Y]
Concept Factorization (CF), as a novel paradigm of representation learning, has demonstrated superior performance in multi-view clustering tasks. It overcomes limitations such as the non-negativity constraint imposed by traditional matrix factorization methods and leverages kernel methods to learn latent representations that capture the underlying structure of the data, thereby improving data representation.
However, existing multi-view concept factorization methods fail to consider the limited labeled information inherent in real-world multi-view data. This often leads to significant performance loss.
To overcome these limitations, we propose a novel semi-supervised multi-view concept factorization model, named SMVCF.
In the SMVCF model, we first extend the conventional single-view CF to a multi-view version, enabling more effective exploration of complementary information across multiple views. We then integrate multi-view CF, label propagation, and manifold learning into a unified framework to leverage and incorporate valuable information present in the data. Additionally, an adaptive weight vector is introduced to balance the importance of different views in the clustering process.
We further develop targeted optimization methods specifically tailored for the SMVCF model. Finally, we conduct extensive experiments on four diverse datasets with varying label ratios to evaluate the performance of SMVCF. The experimental results demonstrate the effectiveness and superiority of our proposed approach in multi-view clustering tasks.

\end{abstract}

\maketitle
\section{Introduction}
\label{intro}

With the rapid growth of data, data sources and features have become increasingly diverse. For example, a news article may be reported by multiple media outlets, a facial image can be captured from different angles, and a web page may contain various elements such as images, text, and hyperlinks. These data, described by different source domains or features, are referred to as multi-view data. Although the data from each view can be used to design single-view representation learning models, this approach fails to leverage the information provided by other views, thereby limiting further improvement in algorithm performance. Therefore, effectively utilizing the information from multi-view data to enhance clustering performance is an important challenge. Multi-view learning has been widely applied in various fields \cite{xie2016unsupervised,xu2016discriminatively,zhao2017multi,liu2016multiple,liu2017spectral,zhang2016multi,yu2022semi}, including computer vision, natural language processing, bioinformatics, and health informatics. 

In the field of multi-view clustering, matrix factorization (MF) methods are commonly employed, particularly Nonnegative Matrix Factorization (NMF) \cite{lee1999learning}. NMF is a dimensionality reduction technique that can extract latent features from high-dimensional multi-view data. By decomposing the original data into a low-rank representation\cite{yu2022graph,yu2021fast}, NMF reduces the dimensionality and captures the underlying structure of the data. This facilitates the clustering process by revealing the intrinsic relationships and patterns among the multi-view samples. Furthermore, Concept Factorization (CF) \cite{xu2004document} as a variant of NMF, inherits all the advantages of NMF and has additional strengths as it can handle both positive and negative values and operate in any data representation space, including kernel feature space.

In recent years, various multi-view clustering methods based on non-negative matrix factorization have been proposed. Liu et al. \cite{liu2013multi} proposed Multi-View Clustering via Joint NMF, which employs a consensus constraint to encourage the coefficient matrices learned from different views to converge towards a consistent consensus matrix. Subsequently, Khan et al. \cite{khan2019weighted} introduced the Weighted Multi-View Data Clustering via Joint NMF method, which incorporates adaptive view weights to enhance the clustering performance.
Wang et al. \cite{wang2016adaptive} presented the Adaptive Multi-View Semi-Supervised NMF, which extends traditional multi-view NMF to the semi-supervised setting by incorporating label information as hard constraints, aiming to achieve better clustering discriminability.
In addition, Wang et al. \cite{wang2017diverse} proposed Diverse NMF, a multi-view clustering method that introduces a diversity term to orthogonalize different data vectors and reduce redundancy in multi-view representations. The accumulated result integrates complementary information from multiple views.
Liu et al. \cite{liu2014partially} introduced Partially Shared Latent Factor Learning (PSLF), a partially shared multi-view learning approach. PSLF assumes that different views share common latent factors while having their specific latent factors. By considering both the consistency and complementarity of multi-view data, PSLF learns a comprehensive partially shared latent representation that enhances clustering discriminability.
Ou et al. \cite{ou2018co} incorporated co-regularization and correlation constraints into multi-view NMF. They leverage the complementarity between different views and propose imposing correlation constraints on the shared latent subspace to obtain shared latent representations when a particular view is corrupted by noise. This approach demonstrates good performance in handling noisy multi-view data.

Based on manifold learning \cite{seung2000manifold}, Cai et al. \cite{cai2010graph} proposed a graph-constrained non-negative matrix factorization model, which highlights the importance of considering manifold learning. Since then, this approach has been widely applied in the field of multi-view learning.
Zhang et al. \cite{zhang2018multi} proposed Graph-Regularized NMF, a multi-view clustering method that incorporates graph regularization and orthogonal constraints. The orthogonal constraints help eliminate relatively less important features, while the graph regularization learns more relevant local geometric structures.
Liang et al. \cite{liang2020semi} introduced a graph-regularized partially shared multi-view NMF method. Building upon the PSLF model \cite{liu2014partially}, this approach incorporates manifold learning and constructs affinity graphs for each view to approximate the geometric structure information in the data.

As a variant of NMF, CF inherits all the advantages of NMF and has additional strengths, making CF a natural choice for the multi-view domain. Wang et al. \cite{wang2016multi} first introduced the Multi-View CF method, extending the traditional single-view CF methods to the multi-view scenario.
Subsequently, Zhan et al. \cite{zhan2018adaptive} proposed the Adaptive Multi-View CF method, which utilizes a view-adaptive weighting strategy to automatically update the weights for each view, further enhancing the ability of CF to handle multi-view problems.

However, the aforementioned studies on multi-view concept factorization overlook an important aspect, which is the presence of a small amount of labeled information in real-world multi-view data. Leveraging the available labeled information can significantly enhance the clustering performance of our model. Therefore, the maximum utilization of limited labeled information becomes a crucial problem to address.

\begin{figure*}[htbp]
	\centering	
	\includegraphics[scale=0.4]{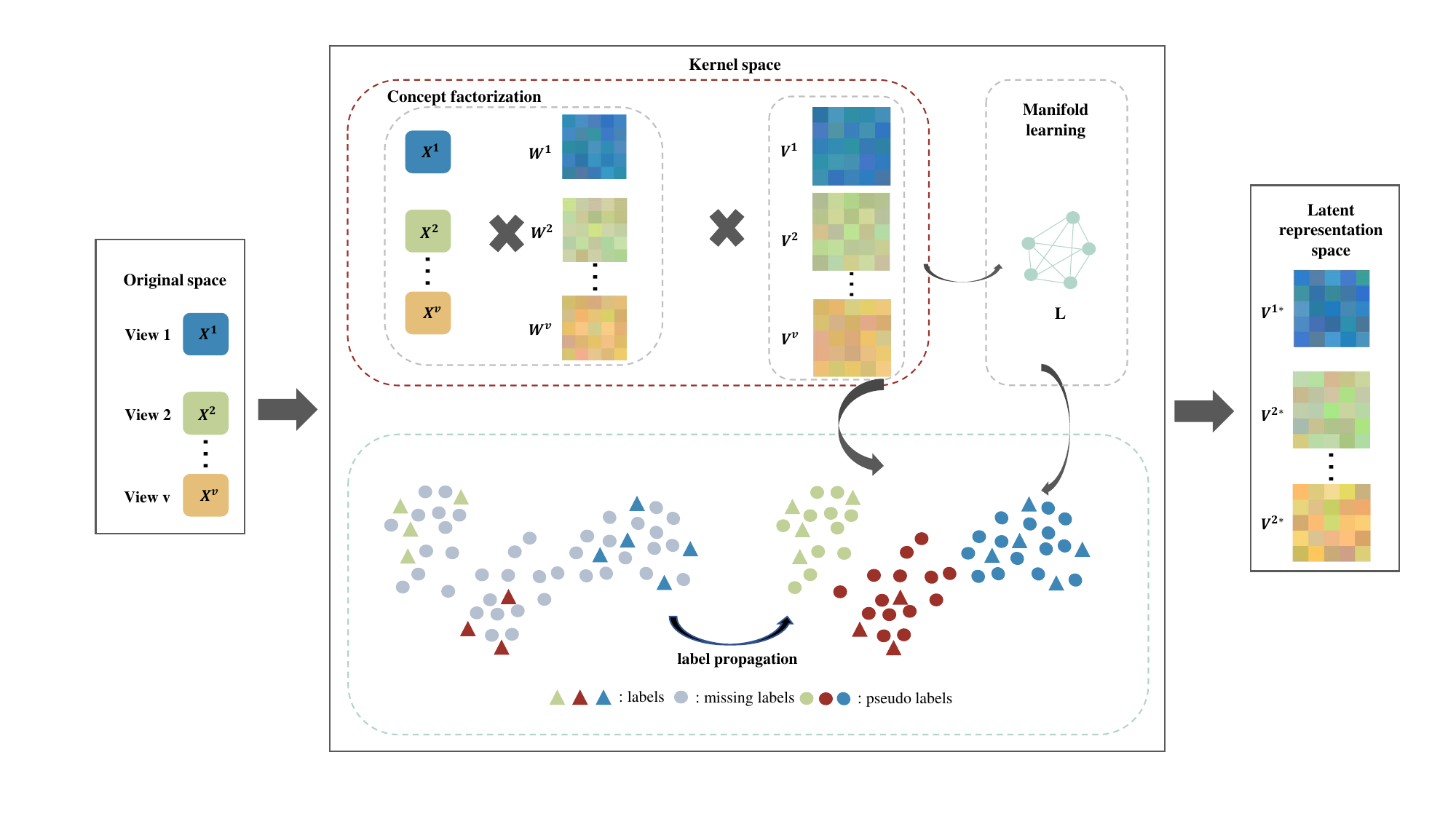}
	\caption{The architecture of the proposed SMVCF}
	\label{Fig:000}
\end{figure*}

To tackle this issue, we propose a novel multi-view concept factorization method called Semi-supervised Multi-View Concept Factorization (SMVCF) model. The framework structure of SMVCF is illustrated in Figure \ref{Fig:000}, and its main contributions can be summarized as follows:

\begin{itemize}
	\item We introduce a new multi-view concept factorization method, SMVCF, which extends the single-view CF to the multi-view scenario and integrates multi-view CF, label propagation, and manifold learning into a unified framework. Moreover, our method combines concept factorization, label propagation, and manifold learning to solve a unified optimization problem.
\end{itemize}

\begin{itemize}
	\item We develop a novel multi-view label learning strategy to utilize the available labeled information. For datasets with a small number of labeled instances, label propagation methods can propagate labels to unlabeled data, thereby improving the model's performance.
\end{itemize}

\begin{itemize}
	\item The proposed SMVCF incorporates an adaptive weight strategy in the learning process to balance the importance of each view, mitigating the adverse effects of information imbalance.
\end{itemize}

\begin{itemize}
	\item We conduct extensive experiments on four different datasets with varying label proportions. The results demonstrate that SMVCF outperforms several state-of-the-art semi-supervised multi-view clustering methods, showcasing its superior performance.
\end{itemize}
\section{Related work}
\subsection{Notations}
For the sake of readability, this section provides a summary of the commonly used mathematical symbols throughout the entire paper in the table\ref{table:21} below.
\begin{table}[htbp]
	\caption{Symbols Commonly Used in This Article}
	\label{table:21}
	\centering
	\scalebox{1.05}{
		\begin{tabular}{cc}
		\hline
		Notations                           & Descriptions                            \\ \hline
		$\mathbb{R}$                                & operational space                       \\
		$\mathbf{X}^{(v)}$                  & The data matrix of the v-th view        \\
		$\mathbf{W}^{(v)}$                  & The basis matrix of the v-th view       \\
		$\mathbf{V}^{(v)}$                  & The coefficient matrix of the v-th view \\
		${\mathbf{L}}$                      & The Laplacian matrix of a data manifold \\
		$\mathbf{D}$                        & The diagonal matrix of a data manifold  \\
		$\mathbf{S}$                        & The weight matrix of a data manifold    \\
	    $\mathbf{B}$                                  & The predicted label matrix              \\
		$\mathbf{Y}$                                  & The ground truth label matrix           \\
		${N}_{p}\left(x_{i}\right)$ & The neighbors of $x_i$                  \\
		${N}_{p}\left(x_{j}\right)$ & The neighbors of $x_j$                  \\
		$\lambda,\beta,\gamma$              & Hyperparameters                         \\
		$\alpha^{v}$                        & The weight matrix of the v-th view      \\
		$\|\cdot\|_{F}$                     & The Frobenius norm                      \\
		$\operatorname{tr}(\cdot)$          & The trace                               \\ \hline
	\end{tabular}
	}
	\centering
\end{table}
\subsection{NMF}
In matrix factorization-based learning methods, Non-negative Matrix Factorization  \cite{lee1999learning} is a technique that approximates a non-negative matrix of sample data by decomposing it into a basis matrix $\mathbf{U}$ and a coefficient matrix $\mathbf{V}$.

Given a data matrix ${\rm{\mathbf{X} = }}\left[ {{x_1},{x_2}, \cdots ,{x_n}} \right] \in {\mathbb{R}^{m \times n}}$, where each column of    $\mathbf{X}$ represents a sample vector, NMF aims to decompose $\mathbf{U}=\left[u_{i k}\right] \in \mathbb{R}^{m \times k}$ and a coefficient matrix $\mathbf{V}=\left[v_{j k}\right] \in \mathbb{R}^{n \times k}$. This can be expressed as follows:

\begin{equation}
	\mathbf{X}\approx \mathbf{U} \mathbf{V}^{\top}.
\end{equation}

Furthermore, we can define the objective function of NMF as follows:
\begin{equation}
	\begin{aligned}
		\mathbf{J}_{\mathbf{N M F}}=&\left\|\mathbf{X}-\mathbf{U V}^{\top}\right\|_{F}^{2} \\	
	\end{aligned}
\end{equation}
\[s.t. \quad \mathbf{U},\mathbf{V} \geq 0.\]

Based on the paper \cite{lee1999learning}, we can derive the update equations for the basis matrix $\mathbf{U}$ and the coefficient matrix $\mathbf{V}$ as follows:
\begin{equation}
	u_{i k}^{t+1}=u_{i k}^{t} \frac{(\mathbf{X V})_{i k}}{\left(\mathbf{U V}^{\top} \mathbf{V}\right)_{i k}}
\end{equation}

\begin{equation}
	v_{j k}^{t+1}=v_{j k}^{t} \frac{\left(\mathbf{X}^{\top} \mathbf{U}\right)_{j k}}{\left(\mathbf{V} \mathbf{U}^{\top}, \mathbf{U}\right)_{j k}}.
\end{equation}
\subsection{CF}
Non-negative Matrix Factorization \cite{lee1999learning} has gained significant attention in the field of clustering \cite{zhou2012fast, zhou2014nonnegative, yu2020graph} over the past few decades. While NMF is effective in extracting latent components from non-negative data, it encounters limitations when applied to real-world data where non-negativity is not always preserved due to noise or outliers. Converting data to non-negative form may disrupt the linear relationships among the data. Additionally, NMF cannot be easily kernelized using kernel methods \cite{muller2001introduction}, as many kernel methods are not applicable to NMF.

To overcome these drawbacks, Xu et al. \cite{xu2004document} proposed the concept factorization approach as an alternative to NMF. CF not only eliminates the constraint of non-negativity but also leverages kernel methods to learn the latent representation of data. By incorporating kernelization, CF can capture nonlinear relationships in the data, which enhances its flexibility compared to NMF. It is worth noting that concept decomposition-based \cite{cai2010locally,shu2018multiple,pei2016concept,shu2019concept,jiang2021semi,liu2011locality,li2015graph,zhang2019robust,li2012clustering,li2015graph}  methods have demonstrated superior performance in handling problems across various domains.

Given a data matrix ${\rm{\mathbf{X} = }}\left[ {{x_1},{x_2}, \cdots ,{x_n}} \right] \in {\mathbb{R}^{m \times n}}$, where ${x_i}$ represents the $i$-th $m$-dimensional feature vector of the data samples, each basis vector ${u_j}$ can be represented as a linear combination of the data samples: ${u_j} = \sum\nolimits_i {{w_{ij}}} {x_i}$,  where ${w_{ij}} \ge 0$. Let  $\mathbf{W} = \left[ {{w_{ij}}} \right] \in {\mathbb{R}^{n \times c}}$. The objective of CF is to find an approximation as follows:

\begin{equation}
	\mathbf{X}\approx \mathbf{X} \mathbf{W} \mathbf{V}^{\top}.
\end{equation}

To measure the reconstruction error, the objective function of CF can be rewritten as follows:
\begin{equation} \label{cf}
	\mathbf{J}_{\mathbf{C F}}=\left\|\mathbf{X}-\mathbf{X} \mathbf{W} \mathbf{V}^{\top}\right\|_{F}^{2}\end{equation}
\[s.t. \quad \mathbf{W}\geq 0, \mathbf{V} \geq 0.\] 

According to the paper \cite{xu2004document}, we can obtain the update rules for problem \eqref{cf} as follows:

\begin{equation}	
	{w}_{i j}^{t+1} \leftarrow w_{i k}^{t} \frac{(\mathbf{K} \mathbf{V})_{i k}}{\left(\mathbf{K} \mathbf{W} \mathbf{V}^{\top} \mathbf{v}\right)_{i k}}
\end{equation}

\begin{equation}
	v_{j k}^{t+1} \leftarrow v_{j k}^{t} \frac{(\mathbf{K} \mathbf{W})_{j k}}{\left(\mathbf{V} \mathbf{W}^{\top} \mathbf{K} \mathbf{W}\right)_{j k}},
\end{equation}
where  $\mathbf{K} = {\mathbf{X}^{\top}\mathbf{X}}$ $\in {\mathbb{R}^{n \times n}}$.

These update rules only involve the inner product of $\mathbf{X}$. However, it is possible to incorporate a kernel function into the matrix to introduce nonlinearity. A detailed explanation can be found in the paper \cite{weinberger2004learning}.
\subsection{Multi-view Clustering}
The general expression of matrix factorization-based multi-view models is given as follows:
\begin{equation}
	\begin{aligned}	
		\min _{\mathbf{U}^{(v)}, \mathbf{V}^{(v)}}& \sum_{v=1}^{m}\left\|\mathbf{X}^{(v)}-\mathbf{U}^{(v)} \mathbf{V}^{(v)}\right\|_{F}^{2}+\Psi\left(\mathbf{V}^{(v)}, \mathbf{V}^{*}\right) 	
	\end{aligned}
\end{equation}
 \[s.t. \quad \mathbf{U}^{(v)}, \mathbf{V}^{(v)}, \mathbf{V}^{*} \geq 0,\]
Where $\mathbf{X}^{(v)}$ represents the data matrix of the $v$-th view. $\mathbf{U}^{(v)}$ is the basis matrix for the $v$-th view, and $\mathbf{V}^{(v)}$ represents the coefficient matrix for the $v$-th view. $\Psi(\cdot)$ is a function that combines different $\mathbf{V}^{(v)}$ matrices to obtain a consistent consensus matrix $\mathbf{V}^{*}$.
\section{Semi-supervised Multi-view Concept Decomposition}

\subsection{Label Propagation}
In the context of multi-view learning, datasets often contain partial label information, and effectively leveraging this limited label information becomes crucial for improving algorithm performance. Label propagation techniques have been widely demonstrated to be effective in previous research \cite{zhou2003learning, wang2006label, tian2012global}. Compared to traditional label learning methods, label propagation methods have several advantages. Firstly, label propagation methods can leverage a large amount of unlabeled data for learning without requiring additional manual labeling costs. Secondly, for datasets with a small number of labels, label propagation methods can propagate label information from known labeled samples to unknown labeled samples through similarity propagation in the data space, thereby improving model performance. Lastly, label propagation methods exhibit high flexibility and robustness, being able to adapt to various data types and task types while being less susceptible to noise and outlier data.

We can establish an undirected graph $\mathbf{G}(V,E)$ and a weight matrix $\mathbf{S}=\left\{s(i,j),i,j = 1,2,\cdots, I_{N}\right\}$ to describe the neighboring relationships between samples. The weight matrix $\mathbf{S}$ can be constructed in the following way:
\begin{equation}
	\begin{small}
s(i, j)=\left\{\begin{array}{ll}\mathrm{e}^{-\frac{\left\|X_{i}-{X}_{j}\right\|_{F}^{2}}{\sigma^{2}}}, & \text { if } {X}_{i} \in {N}_{p}\left({X}_{j}\right) \text { and } {X}_{j} \in {N}_{p}\left({X}_{i}\right) \\ 0, & \text { otherwise }\end{array}\right.
	\end{small}
\end{equation}
Where ${N}_{p}\left({X}_{i}\right)$ and ${N}_{p}\left({X}_{j}\right)$ denote the sets of $p$ nearest samples to ${X}_{i}$ and ${X}_{j}$ in the graph $G$, respectively. $\sigma$ is a hyperparameter.

Previous studies have shown \cite{tian2012global} that samples that are close in the sample space should have the same label. Therefore, if the dataset consists of ${I}_{N}$ samples and contains label information, the label propagation problem can be rewritten as follows:

\begin{equation} \label{yc}
	\begin{aligned}	
		 \min &\sum_{i=1}^{I_{N}} \sum_{j=1}^{I_{N}}\|\mathbf{B}(i,:)-\mathbf{B}(j,:)\|_{2}^{2} s(i, j)
		\\ & +\sum_{i=1}^{I_{N}}\|\mathbf{B}(i,:)-\mathbf{Y}(i,:)\|_{2}^{2} a(i, i),
	\end{aligned}
\end{equation}

Where $\mathbf{B} \in \mathbb{R}^{I_{N} \times k}$ is the predicted label matrix, $\mathbf{Y} \in \mathbb{R}^{I_{N} \times k}$ is the true label matrix, and $\mathbf{Y}(i,:)=[0,0, \cdots, 1, \cdots, \\ 0,0]^{\top} $ $\mathbb \in{R}^{1 \times k}$. Here, $k$ represents the number of classes for the samples. $\mathbf{A}=\left\{a(i, i), i, j \in 1,2, \cdots, I_{N}\right\} \in \mathbb{R}^{I_{N} \times I_{N}}$ denotes the diagonal indicator matrix.
\begin{equation}
	a(i, i)=\left\{\begin{array}{l}1,\; \text{if}\; {X}_{i}\;   \text{labeled.} \\ 0, \text{ otherwise. }\end{array}\right.
\end{equation}

When given labeled samples ${X}_{i}$ and unlabeled samples ${X}_{j}$, if $s(i, j)$ is sufficiently large, minimizing \eqref{yc} ensures that the predicted label $\mathbf{B}(j,:)$ for sample ${X}_{j}$ will be very close to the true label $\mathbf{Y}(i,:)$ of sample ${X}_{i}$.

\subsection{Objective Function of SMVCF}
We integrate multi-view CF, label propagation, and manifold learning into a unified framework and propose a semi-supervised multi-view concept factorization model. Given a dataset with $n_v$ views $\left\{\mathbf{X}^{(v)}\right\}_{v=1}^{n_v}$, where $\mathbf{X}^{(v)}=\left[x_{1}^{(v)}, x_{2}^{(v)}, \cdots, x_{n}^{(v)}\right]$ is the input matrix of the $v$-th view. It is important to note that a good low-dimensional representation vector $\mathbf{V}^{(v)}(i,:)$ should ideally have a small Euclidean distance to its corresponding label vector, resulting in better discriminative power. Therefore, the objective function of our SMVCF model is formulated as follows:
\begin{equation}\label{SMVCF1}
	\begin{aligned} \min _{\mathbf{W}^{(v)}, \mathbf{V}^{(v)}, \alpha^{v}}& \sum_{v=1}^{n_{v}} \alpha^{v}  \left(\left\|\mathbf{X}^{(v)}-\mathbf{X}^{(v)} \mathbf{W}^{(v)} \left(\mathbf{V}^{(v)}\right)^{\top}\right\|_{\mathrm{F}}^{2}\right. \\ & +\lambda \sum_{i=1}^{n_{v}} \sum_{j=1}^{n_{v}}\left\|\mathbf{V}^{(v)}(i,:)-\mathbf{V}^{(v)}(j,:)\right\|_{2}^{2} s(i, j) \\ & \left.+\beta \sum_{i=1}^{n_{v}}\left\|\mathbf{Y}^{(v)}(i,:)-\mathbf{V}^{(v)}(i,:)\right\|_{2}^{2} a(i, i)\right)			
\end{aligned}
\end{equation}
\[s.t. \quad \forall v, \mathbf{W}^{(v)} \geq 0, \mathbf{V}^{(v)} \geq 0, \alpha^{v} \geq 0, \sum_{v=1}^{n_{v}} \alpha^{v}=1,\]
Where $\mathbf{V}^{(v)}(i,:)$ and $\mathbf{V}^{(v)}(j,:)$ represent the $i$-th and $j$-th rows of the factor matrix $\mathbf{V}^{(v)}$ in the $v$-th view. Equation \eqref{SMVCF1} can also be rewritten in the following form:
\begin{equation}\label{SMVCF2}
	\begin{aligned}
		\min _{\mathbf{W}^{(v)}, \mathbf{V}^{(v)}, \alpha^{v}} \sum_{v=1}^{n_{v}}& \alpha^{v}  \left(\left\|\mathbf{X}^{(v)}-\mathbf{X}^{(v)} \mathbf{W}^{(v)} \left(\mathbf{V}^{(v)}\right)^{\top}\right\|_{\mathrm{F}}^{2}
	\right.	\\ & +\lambda \operatorname{Tr}\left(\mathbf{V}^{(v)^{\top}} \mathbf{L}^{(v)} \mathbf{V}^{(v)}\right)  \\ & \left.+\beta \operatorname{Tr}\left(\left(\mathbf{V}^{(v)}-\mathbf{Y}^{(v)}\right)^{\top} \mathbf{A}^{(v)}\left(\mathbf{V}^{(v)}-\mathbf{Y}^{(v)}\right)\right)\right) 	 
	\end{aligned}
\end{equation}
\[s.t. \quad \forall v, \mathbf{W}^{(v)} \geq 0, \mathbf{V}^{(v)} \geq 0, \alpha^{v} \geq 0, \sum_{v=1}^{n_{v}} \alpha^{v}=1.\]

However, it is important to note that when one of the views has a weight of 1 and the weights of other views are all 0, Equation \eqref{SMVCF1} will have an invalid solution in terms of $\alpha^{v}$. However, if we solve Equation \eqref{SMVCF2}:
\begin{equation}
	\begin{aligned}
		&\min _{\alpha^{v}} \sum_{v=1}^{n_{v}}\left(\alpha^{v}\right)^{2}
		\\  &s.t. \forall v, \alpha^{v} \geq 0, \sum_{v=1}^{n_{v}} \alpha^{v}=1,
	\end{aligned}
\end{equation}

The optimal solution is for all views to have equal weights: $\frac{1}{n^{v}}$. By combining Equations \eqref{SMVCF1} and \eqref{SMVCF2}, we can avoid the occurrence of invalid solutions. In summary, the final objective function can be formulated as follows:
\begin{equation}\label{SMVCF3}
	\begin{aligned}
		\min _{\mathbf{W}^{(v)}, \mathbf{V}^{(v)}, \alpha^{v}} \sum_{v=1}^{n_{v}}& \alpha^{v}  \left(\left\|\mathbf{X}^{(v)}-\mathbf{X}^{(v)} \mathbf{W}^{(v)} \left(\mathbf{V}^{(v)}\right)^{\top}\right\|_{\mathrm{F}}^{2}\right.
	\\&	+\lambda \operatorname{Tr}\left(\mathbf{V}^{(v)^{\top}} \mathbf{L}^{(v)} \mathbf{V}^{(v)}\right)+\gamma \sum_{v=1}^{n_{v}}\left(\alpha^{v}\right)^{2} \\ & \left.+\beta \operatorname{Tr}\left(\left(\mathbf{V}^{(v)}-\mathbf{Y}^{(v)}\right)^{\top} \mathbf{A}^{(v)}\left(\mathbf{V}^{(v)}-\mathbf{Y}^{(v)}\right)\right)\right)  		
	\end{aligned}
\end{equation}
\[ s.t. \quad \forall v, \mathbf{W}^{(v)} \geq 0, \mathbf{V}^{(v)} \geq 0, \alpha^{v} \geq 0, \sum_{v=1}^{n_{v}} \alpha^{v}=1.\]
The diagonal matrix can be represented as 
$\mathbf{D}=\left\{d(i, i) \right. \\ \left. =\sum_{j=1}^{n_v} s(i, j), i, j \in 1,2, \cdots, n_v\right.\} \in \mathbb{R}^{n_v \times n_v}$. The Laplacian matrix is defined as $\mathbf{L}=\mathbf{D}-\mathbf{S}$. In Equation \eqref{SMVCF3}, $\lambda, \beta, \gamma$ are hyperparameters.
\subsection{Optimization of SMVCF Problem}
We have designed an iterative update algorithm to solve problem \eqref{SMVCF3}. This iterative update algorithm can be roughly divided into three steps:1) Fix $\mathbf{V}^{(v)}$ and $\alpha^{v}$, update $\mathbf{W}^{(v)}$;  2) Fix $\mathbf{W}^{(v)}$ and $\alpha^{v}$, update $\mathbf{V}^{(v)}$; 3) Fix $\mathbf{W}^{(v)}$ and $\mathbf{V}^{(v)}$, update $\alpha^{v}$.

1) Fix $\mathbf{V}^{(v)}$ and $\alpha^{v}$, update$\mathbf{W}^{(v)}$.
	\begin{equation} \label{O1}
		\begin{aligned}
			\mathcal{O_1}=\min _{\mathbf{W}^{(v)}, \mathbf{V}^{(v)}}& \|\mathbf{X}^{(v)}-\mathbf{X}^{(v)} \mathbf{W}^{(v)} \left(\mathbf{V}^{(v)}\right)^{\top} \|_{\mathrm{F}}^{2}
		\end{aligned}
	\end{equation}
\[ s.t.  \quad \forall v, \mathbf{W}^{(v)} \geq 0, \mathbf{V}^{(v)} \geq 0.\]

By defining $\mathbf{K}^{(v)}=\left(\mathbf{X}^{(v)}\right)^{\top} \mathbf{X}^{(v)}$, equation \eqref{O1} can be rewritten as:
\begin{equation}
	\begin{aligned}
	\begin{array}{l}\left\|\mathbf{X}^{(v)}-\mathbf{X}^{(v)} \mathbf{W}^{(v)} \left(\mathbf{V}^{(v)}\right)^{\top}\right\|_{\mathrm{F}}^{2} 
		\\  =  \operatorname{Tr}\left(\mathbf{X}^{(v)}-\mathbf{X}^{(v)} \mathbf{W}^{(v)} \left(\mathbf{V}^{(v)}\right)^{\top}\right)^{\top} \\ \left(\mathbf{X}^{(v)}-\mathbf{X}^{(v)} \mathbf{W}^{(v)} \left(\mathbf{V}^{(v)}\right)^{\top}\right) \\  = \operatorname{Tr}\left(\mathbf{I}-\mathbf{W}^{(v)} \left(\mathbf{V}^{(v)}\right)^{\top}\right)^{\top} \mathbf{K}^{(v)} \left(\mathbf{I}-\mathbf{W}^{(v)} \left(\mathbf{V}^{(v)}\right)^{\top}\right) \\ = \operatorname{Tr}\left(\mathbf{K}^{(v)}-2\mathbf{V}^{(v)}\left(\mathbf{W}^{(v)}\right)^{\top} \mathbf{K}^{(v)}
		\right.\\ \left. +\mathbf{V}^{(v)}\left(\mathbf{W}^{(v)}\right)^{\top} \mathbf{K}^{(v)} \mathbf{W}^{(v)}\left(\mathbf{V}^{(v)}\right)^{\top} \right).\end{array}
	\end{aligned}
\end{equation}

Let $\Psi^{(v)}  = [{\psi _{ik}^{(v)}}]$ be the Lagrange multipliers for $\mathbf{W}^{(v)} \geq 0$, then we can obtain the Lagrangian equation $\mathcal{L}_1$:
\begin{equation}
	\begin{array}{c}\mathcal{L}_{1}=\operatorname{Tr}\left(\mathbf{K}^{(v)}-2 \mathbf{V}^{(v)}\left(\mathbf{W}^{(v)}\right)^{\top} \mathbf{K}^{(v)}+\Psi^{(v)}\left(\mathbf{W}^{(v)}\right)^{\top}\right. \\ \left.+\mathbf{V}^{(v)}\left(\mathbf{W}^{(v)}\right)^{\top} \mathbf{K}^{(v)} \mathbf{W}^{(v)}\left(\mathbf{V}^{(v)}\right)^{\top}\right).\end{array}
\end{equation}

Taking the first-order partial derivative of $\mathcal{L}_1$ with respect to $\mathbf{W}^{(v)}$ yields:
\begin{equation}	
	\frac{{\partial {\rm{\mathcal{L}_1}}}}{{\partial {\rm{\mathbf{W}^{(v)}}}}} =  -2\mathbf{K}^{(v)}\mathbf{V}^{(v)} +2\mathbf{K}^{(v)}\mathbf{W}^{(v)}\left(\mathbf{V}^{(v)}\right)^{\top}\mathbf{V}^{(v)}   +\Psi^{(v)}.	
\end{equation}

By applying the KKT (Karush-Kuhn-Tucker) conditions, ${\psi _{ik}^{(v)}}{w_{ik}^{(v)}} = 0$, we can obtain:
\begin{equation}
	\left(-\mathbf{K}^{(v)}\mathbf{V}^{(v)}+ \mathbf{K}^{(v)}\mathbf{W}^{(v)}\left(\mathbf{V}^{(v)}\right)^{\top}\mathbf{V}^{(v)}  \right)_{ik}{w_{ik}^{(v)}}= 0.
\end{equation}

Therefore, we can obtain the update rule for $w_{ik}^{(v)}$ as follows:
\begin{equation}
	w_{ik}^{(v)} \leftarrow w_{ik}^{(v)} \frac{\left(\mathbf{K}^{(v)}\mathbf{V}^{(v)}\right)_{i k}}{\left(\mathbf{K}^{(v)}\mathbf{W}^{(v)}\left(\mathbf{V}^{(v)}\right)^{\top}\mathbf{V}^{(v)} \right)_{i k}}.
\end{equation}

2) Fix $\mathbf{W}^{(v)}$ and $\alpha^{v}$, update$\mathbf{V}^{(v)}$.
\begin{equation}\label{O2}
	\begin{aligned}
		\mathcal{O_2} =	\min _{\mathbf{W}^{(v)}, \mathbf{V}^{(v)}} & \left\|\mathbf{X}^{(v)}-\mathbf{X}^{(v)} \mathbf{W}^{(v)}\left(\mathbf{V}^{(v)}\right)^{\top}\right\|_{\mathrm{F}}^{2}
		 \\& +\lambda \operatorname{Tr}\left(\mathbf{V}^{(v)^{\top}} \mathbf{L}^{(v)} \mathbf{V}^{(v)}\right) \\ & +\beta \operatorname{Tr}\left(\left(\mathbf{V}^{(v)}-\mathbf{Y}^{(v)}\right)^{\top} \mathbf{A}^{(v)}\left(\mathbf{V}^{(v)}-\mathbf{Y}^{(v)}\right)\right) 	
	\end{aligned}
\end{equation}
\[ s.t.\quad \forall v, \mathbf{W}^{(v)} \geq 0, \mathbf{V}^{(v)} \geq 0.\]

By defining $\mathbf{K}^{(v)}=\left(\mathbf{X}^{(v)}\right)^{\top} \mathbf{X}^{(v)}$, equation \eqref{O2} can be rewritten as:	
\begin{equation}
	\begin{aligned}
		\begin{array}{l}\left\|\mathbf{X}^{(v)}-\mathbf{X}^{(v)} \mathbf{W}^{(v)}\left(\mathbf{V}^{(v)}\right)^{\top}\right\|_{\mathrm{F}}^{2}+\lambda \operatorname{Tr}\left(\mathbf{V}^{(v)^{\top}} \mathbf{L}^{(v)} \mathbf{V}^{(v)}\right) \\ +\beta \operatorname{Tr}\left(\left(\mathbf{V}^{(v)}-\mathbf{Y}^{(v)}\right)^{\top} \mathbf{A}^{(v)}\left(\mathbf{V}^{(v)}-\mathbf{Y}^{(v)}\right)\right)\\ = 
			\operatorname{Tr}\left(\mathbf{K}^{(v)}-2\mathbf{V}^{(v)}\left(\mathbf{W}^{(v)}\right)^{\top} \mathbf{K}^{(v)}
			\right. \\ \left. +\mathbf{V}^{(v)}\left(\mathbf{W}^{(v)}\right)^{\top} \mathbf{K}^{(v)} \mathbf{W}^{(v)}\left(\mathbf{V}^{(v)}\right)^{\top}\right) +\lambda \operatorname{Tr}\left(\mathbf{V}^{(v)^{\top}} \mathbf{L}^{(v)} \mathbf{V}^{(v)}\right) \\ +\beta \operatorname{Tr}\mathbf{A}^{(v)}\left(  \left(\mathbf{V}^{(v)}\right)^{\top} \mathbf{V}^{(v)}-\left(\mathbf{V}^{(v)}\right)^{\top} \mathbf{Y}^{(v)}-\left(\mathbf{Y}^{(v)}\right)^{\top} \mathbf{V}^{(v)} \right. 
			\\ \left.-\left(\mathbf{Y}^{(v)}\right)^{\top} \mathbf{Y}^{(v)}\right) .    			
		\end{array}
	\end{aligned}
\end{equation}

Let $\Phi^{(v)} = [{\varphi _{jk}^{(v)}}]$ be the Lagrange multipliers for $\mathbf{V}^{(v)} \geq 0$. Then we can obtain the Lagrange equation$\mathcal{L}_2$:
\begin{equation}	
	\begin{aligned}
		\mathcal{L}_2 =& \operatorname{Tr}\left(\mathbf{K}^{(v)}-2\mathbf{V}^{(v)}\left(\mathbf{W}^{(v)}\right)^{\top} \mathbf{K}^{(v)} \right. \\& \left.+\mathbf{V}^{(v)}\left(\mathbf{W}^{(v)}\right)^{\top} \mathbf{K}^{(v)} \mathbf{W}^{(v)}\left(\mathbf{V}^{(v)}\right)^{\top}\right) 
		\\ &+\beta \operatorname{Tr}\mathbf{A}^{(v)}\left(  \left(\mathbf{V}^{(v)}\right)^{\top} \mathbf{V}^{(v)}-\left(\mathbf{V}^{(v)}\right)^{\top} \mathbf{Y}^{(v)}
		\right. \\ & \left.-\left(\mathbf{Y}^{(v)}\right)^{\top} \mathbf{V}^{(v)}-\left(\mathbf{Y}^{(v)}\right)^{\top} \mathbf{Y}^{(v)}\right)\\ & +\lambda \operatorname{Tr}\left(\mathbf{V}^{(v)^{\top}} \mathbf{L}^{(v)} \mathbf{V}^{(v)}\right) +\operatorname{Tr}\left(\Phi^{(v)} \left(\mathbf{V}^{(v)}\right)^{\top}\right).
	\end{aligned}
\end{equation}

To solve for the first-order partial derivative of $\mathcal{L}_2$ with respect to $\mathbf{V}^{(v)}$, we obtain:
\begin{equation}
	\begin{aligned}
		\frac{{\partial {\rm{\mathcal{L}_2}}}}{{\partial {\rm{\mathbf{V}^{(v)}}}}} = & -2\mathbf{K}^{(v)}\mathbf{W}^{(v)} +2\mathbf{K}^{(v)}\mathbf{V}^{(v)}\left(\mathbf{W}^{(v)}\right)^{\top}\mathbf{W}^{(v)} \\ & +2\lambda \mathbf{L}^{(v)} \mathbf{V}^{(v)}  +2\beta \left(\mathbf{A}^{(v)}\mathbf{V}^{(v)} - \mathbf{A}^{(v)}\mathbf{Y}^{(v)}\right) +\Phi^{(v)}.			
	\end{aligned}	
\end{equation}

By applying the KKT conditions, ${\phi _{jk}^{(v)}}{v_{jk}^{(v)}} = 0$, we can obtain:
\begin{equation}\label{O22}
	\begin{array}{l}\left(-\mathbf{K}^{(v)} \mathbf{W}^{(v)}+\mathbf{K}^{(v)} \mathbf{V}^{(v)}\left(\mathbf{W}^{(v)}\right)^{\top} \mathbf{W}^{(v)}\right. \\ \left.+\lambda \mathbf{L}^{(v)} \mathbf{V}^{(v)}+\beta\left(\mathbf{A}^{(v)} \mathbf{V}^{(v)}-\mathbf{A}^{(v)} \mathbf{Y}^{(v)}\right)\right)_{j k} v_{j k}^{(v)}=0,\end{array}	
\end{equation}
Where $ \mathbf{L} = \mathbf{D} - \mathbf{S} $, equation \eqref{O22} can be rewritten as:
\begin{equation}
	\begin{array}{l}\left(-\mathbf{K}^{(v)} \mathbf{W}^{(v)}+\mathbf{K}^{(v)} \mathbf{V}^{(v)}\left(\mathbf{W}^{(v)}\right)^{\top} \mathbf{W}^{(v)}+\lambda \mathbf{D}^{(v)} \mathbf{V}^{(v)} \right. \\ \left. -\lambda \mathbf{S}^{(v)} \mathbf{V}^{(v)} +\beta\left(\mathbf{A}^{(v)} \mathbf{V}^{(v)}-\mathbf{A}^{(v)} \mathbf{Y}^{(v)}\right)\right)_{j k} v_{j k}^{(v)}=0.\end{array}	
\end{equation}

we can derive the update rule for $v_{jk}^{(v)}$ as follows:
\begin{small}
\begin{equation}
	v_{jk}^{(v)} \leftarrow v_{jk}^{(v)} \frac{\left(\mathbf{K}^{(v)}\mathbf{W}^{(v)} +\lambda \mathbf{S}^{(v)}\mathbf{V}^{(v)} +\beta \mathbf{A}^{(v)}\mathbf{Y}^{(v)} \right)_{j k}}{\left(\mathbf{K}^{(v)}\mathbf{V}^{(v)}\left(\mathbf{W}^{(v)}\right)^{\top}\mathbf{W}^{(v)} \\ +\lambda \mathbf{D}^{(v)}\mathbf{V}^{(v)} +\beta\mathbf{A}^{(v)}\mathbf{V}^{(v)}\right)_{j k}}.			
\end{equation}
	\end{small}

3) Fix $\mathbf{W}^{(v)}$ and $\mathbf{V}^{(v)}$, update $\alpha^{v}$.
\begin{equation} \label{O3}
	\begin{aligned}
	\mathcal{O_3} = \min _{ \alpha^{v}} \sum_{v=1}^{n_{v}}& \alpha^{v}  \left(\left\|\mathbf{X}^{(v)}-\mathbf{X}^{(v)} \mathbf{W}^{(v)} \left(\mathbf{V}^{(v)}\right)^{\top}\right\|_{\mathrm{F}}^{2} \right. \\ & \left. +\lambda \operatorname{Tr}\left(\mathbf{V}^{(v)^{\top}} \mathbf{L}^{(v)} \mathbf{V}^{(v)}\right)+\gamma \sum_{v=1}^{n_{v}}\left(\alpha^{v}\right)^{2}\right. \\ & \left.+\beta \operatorname{Tr}\left(\left(\mathbf{V}^{(v)}-\mathbf{Y}^{(v)}\right)^{\top} \mathbf{A}^{(v)}\left(\mathbf{V}^{(v)}-\mathbf{Y}^{(v)}\right)\right)\right)  
\end{aligned}
\end{equation}
\[s.t. \quad \forall v,  \alpha^{v} \geq 0, \sum_{v=1}^{n_{v}} \alpha^{v}=1.\]
	
Let\begin{small}$f^{(v)} = \left\|\mathbf{X}^{(v)}-\mathbf{X}^{(v)} \mathbf{W}^{(v)} \left(\mathbf{V}^{(v)}\right)^{\top}\right\|_{\mathrm{F}}^{2} +\lambda \operatorname{Tr}\left(\mathbf{V}^{(v)^{\top}} \mathbf{L}^{(v)} \mathbf{V}^{(v)}\right)
\\ +\beta \operatorname{Tr}\left(\left(\mathbf{V}^{(v)}-\mathbf{Y}^{(v)}\right)^{\top} \mathbf{A}^{(v)}\left(\mathbf{V}^{(v)}-\mathbf{Y}^{(v)}\right)\right)$\end{small}, equation \eqref{O3} can be rewritten as:
\begin{equation}\label{O33}
	\begin{aligned}
		&\min _{\alpha}\left\|\alpha+\frac{1}{2 \gamma} f\right\|_{2}^{2} 
	\end{aligned}
\end{equation}
\[ s.t. \quad \alpha \geq 0, 1^{\top} \alpha=1,\]
Where $\alpha=\left[\alpha^{1}, \alpha^{2}, \ldots, \alpha^{n_{v}}\right]^{\top}$ and $f=\left[f^{1}, f^{2}, \ldots, f^{n_{v}}\right]^{\top}$. The Lagrangian equation for problem \eqref{O33} is given by:
	\begin{equation}
		\mathcal{L}_3 = \left\|\ {\alpha}+\frac{1}{2 \gamma} f\right\|_{2}^{2}+\rho\left(1-\mathbf{1}^{\top} \ {\alpha}\right)+\ {\zeta}^{\top}(-\ {\alpha}),
	\end{equation}
Where $\rho$ and $\zeta$ are Lagrange multipliers, with $\rho$ being a scalar and $\zeta$ being a column vector. According to the KKT conditions, the optimal solution for $\alpha$ is given by:
\begin{equation}
	\alpha=\left(-\frac{1}{2 \gamma} f+\zeta 1\right)_{+}
\end{equation}
\section{Experiments}
In this section, we compare SMVCF with advanced semi-supervised multi-view algorithms. All experiments were conducted on a PC with an Intel i5 9500T CPU and 16GB of RAM.
\subsection{Datasets}
The detailed information about the datasets used in this experiment is summarized in Table \ref{table:41}.
\begin{itemize}
	\item [(1)]
	$\mathbf{NGs:}$ The $20Newsgroups$ dataset consists of news articles categorized into 20 topics. $NGs$ is a subset of the $20Newsgroups$ dataset, comprising 500 news articles. The dataset is divided into three views based on three preprocessing methods, and for detailed preprocessing steps, please refer to the reference \cite{hussain2010improved}. Each view has the same dimensionality, with $\mathbf{X} \in \mathbb{R}^{2000 \times 500}$.
\end{itemize}
\begin{itemize}
	\item [(2)]
	$\mathbf{BBCSport:}$ The $BBCSport$ dataset consists of 544 news articles from five different sports categories. Each news article in the dataset has two views, where $\mathbf{X}^{(1)} \in \mathbb{R}^{3183 \times 544}$ and $\mathbf{X}^{(2)} \in \mathbb{R}^{3203 \times 544}$.
\end{itemize}
\begin{itemize}
	\item [(3)]
	$\mathbf{BBC:}$ The $BBC$ dataset consists of 685 news articles from five different topic domains. Each news article in the dataset has four views, where $\mathbf{X}^{(1)} \in \mathbb{R}^{4659 \times 685}$, $\mathbf{X}^{(2)} \in \mathbb{R}^{4633 \times 685}$, $\mathbf{X}^{(3)} \in \mathbb{R}^{4655 \times 685}$, and $\mathbf{X}^{(4)} \in \mathbb{R}^{4684 \times 685}$.
\end{itemize}
\begin{itemize}
	\item [(4)]
	$ \mathbf{3Sources:}$ The $3Sources$ dataset consists of 169 news articles from six different topic domains. All the news articles are reported by three news agencies: The Guardian, Reuters, and BBC. Each news agency corresponds to one view, where $\mathbf{X}^{(1)} \in \mathbb{R}^{3560 \times 169}$, $\mathbf{X}^{(2)} \in \mathbb{R}^{3631 \times 169}$, and $\mathbf{X}^{(3)} \in \mathbb{R}^{3068 \times 169}$.
\end{itemize}
\begin{table}[htbp]	
	\caption{Details of the datasets}	
	\begin{tabular}{cccc}
		\hline
		Datasets      & Instances  & Classes  & Views  \\ \hline
		NGs      & 500    & 5      & 3     \\
		BBCSport & 544    & 5      & 2     \\
		BBC      & 685    & 5      & 4     \\
		3sources & 169    & 6      & 3     \\ \hline
	\end{tabular}
	\label{table:41}	
\end{table}
\subsection{Compared Algorithms}
To evaluate the performance of SMVCF, we compared it with the following four semi-supervised multi-view methods.
\begin{itemize}
	\item $\mathbf{DICS}$ \cite{zhang2018multi}: This is a semi-supervised multi-view learning method based on NMF. The proposed approach aims to explore both discriminative and non-discriminative information present in the common and view-specific components across different views through joint non-negative matrix factorization. It also incorporates graph regularization and orthogonal constraints. The orthogonal constraints help eliminate relatively less important features, while the graph regularization learns more relevant local geometric structures.
\end{itemize}

\begin{itemize}
	\item $\mathbf{PSLF}$ \cite{liu2014partially}: This is a semi-supervised multi-view learning method based on NMF. The approach assumes that different views share common latent factors while also having their specific latent factors. By considering both the consistency and complementarity of multi-view data, PSLF learns a comprehensive partially shared latent representation that enhances clustering discriminability.
\end{itemize}

\begin{itemize}
	\item $\mathbf{GPSNMF}$ \cite{liang2020semi}: This is a semi-supervised multi-view learning method based on NMF. Building upon the PSLF model, GPSNMF incorporates manifold learning and constructs affinity graphs for each view to approximate the geometric structure information in the data. Additionally, an efficient $L_{2,1}$-norm regularized regression matrix is employed to learn from labeled samples.
\end{itemize}

\begin{itemize}
	\item $\mathbf{MVSL}$ \cite{peng2019graph}: This is a novel semi-supervised multi-view semantic subspace learning method based on NMF. The proposed approach achieves joint analysis of multi-view data by sharing semantic subspaces across multiple views. It employs a novel graph regularization approach to preserve the geometric structure of the data and utilizes non-negative matrix factorization to learn the semantic subspaces for each view.
\end{itemize}
\subsection{Parameter Sensitivity}
\begin{figure*}[htbp] 
	\centering	
	\subfigure[\label{Fig:11}]{%
		\includegraphics[scale=0.32]{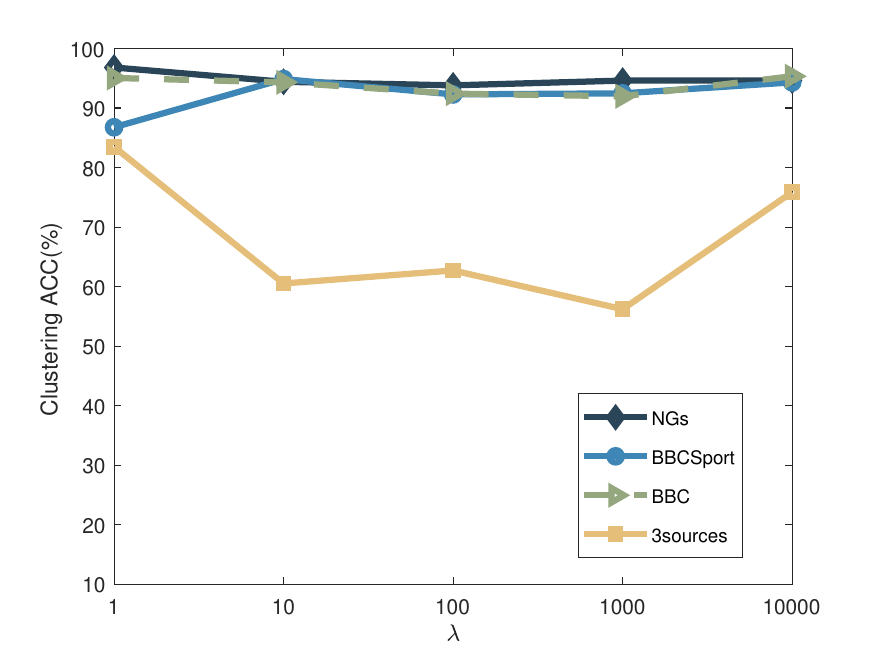}}
	\subfigure[\label{Fig:12}]{%
		\includegraphics[scale=0.32]{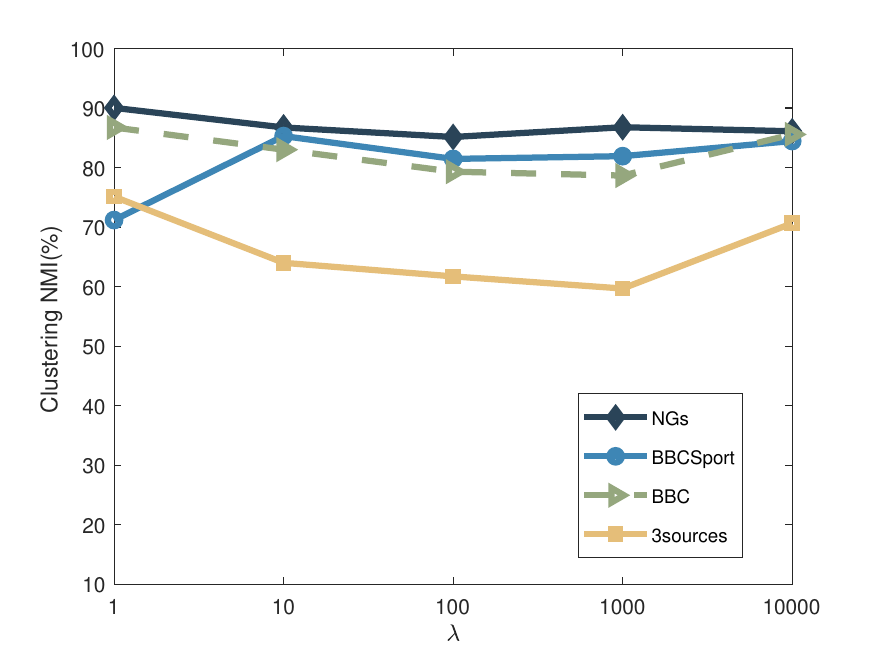}}
	\subfigure[\label{Fig:13}]{%
		\includegraphics[scale=0.32]{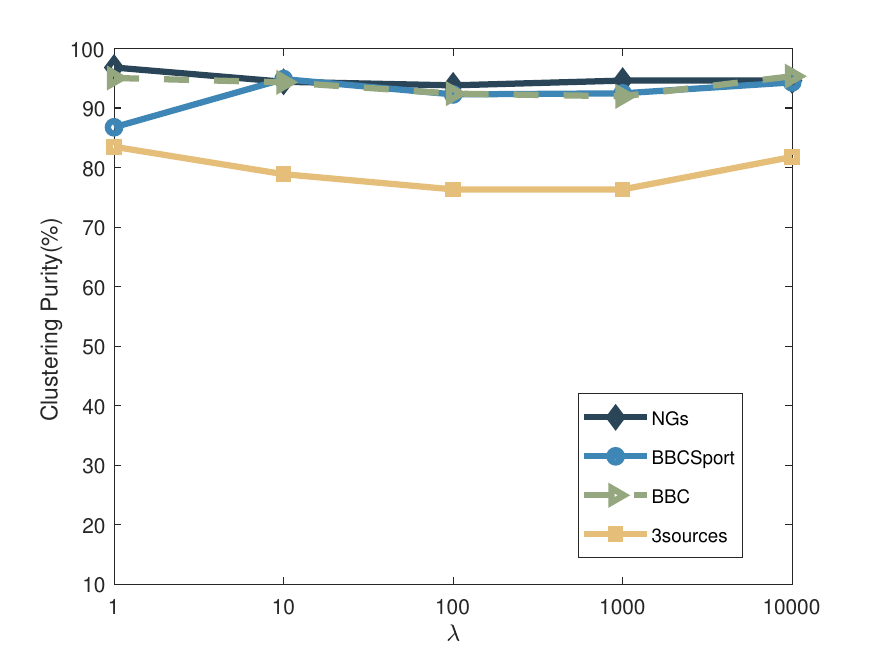}}
	\caption{Clustering ACC$(\%)$, NMI$(\%)$ and Purity$(\%)$ of SMVCF with different $\lambda$ on NGs, BBCSport, BBC, and 3sources datasets}
	\label{Fig:4-1}
\end{figure*}
\begin{figure*}[htbp]
	\centering	
	\subfigure[\label{Fig:14}]{%
		\includegraphics[scale=0.32]{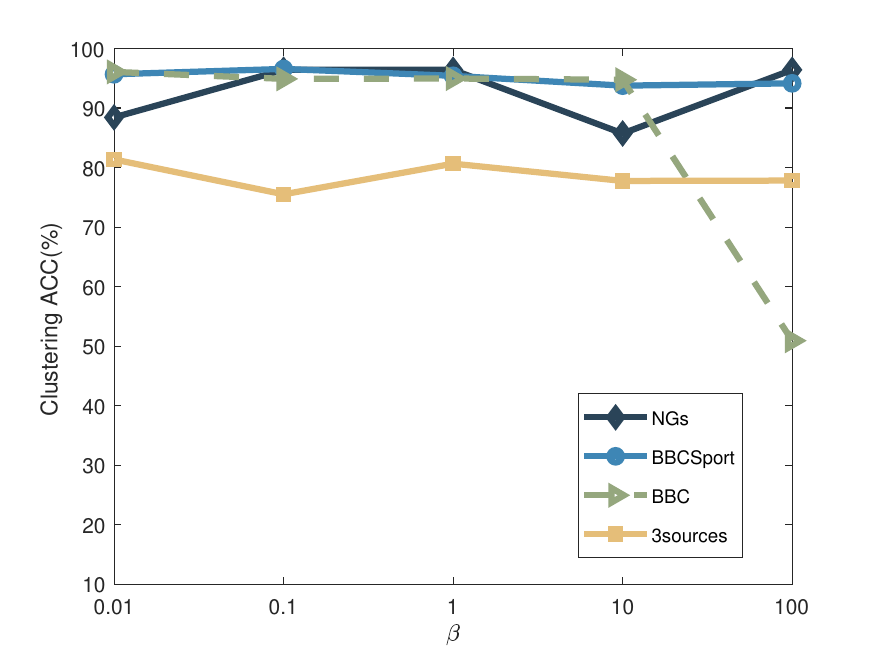}}
	\subfigure[\label{Fig:15}]{%
		\includegraphics[scale=0.32]{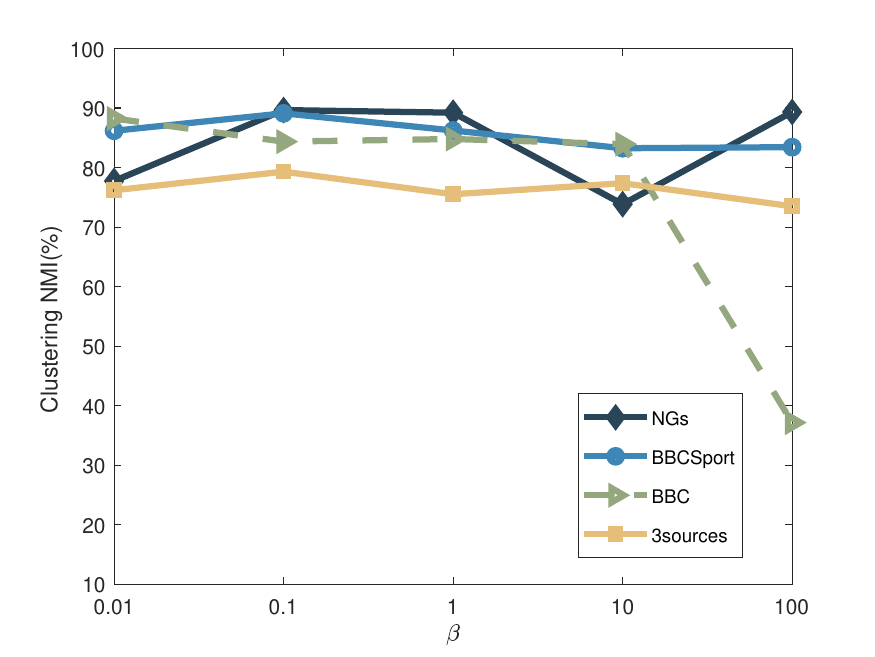}}	
	\subfigure[\label{Fig:16}]{%
		\includegraphics[scale=0.32]{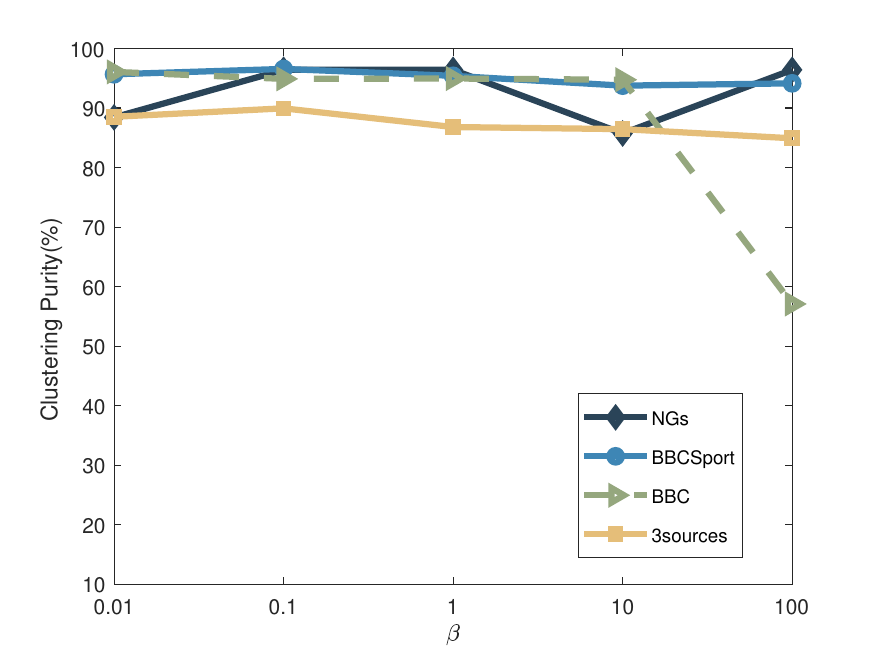}}
	\caption{Clustering ACC$(\%)$, NMI$(\%)$ and Purity$(\%)$ of SMVCF with different $\beta$ on NGs, BBCSport, BBC, and 3sources datasets}
	\label{Fig:4-2}
\end{figure*}
\begin{figure*}[htbp]
	\centering	
	\subfigure[\label{Fig:17}]{%
		\includegraphics[scale=0.32]{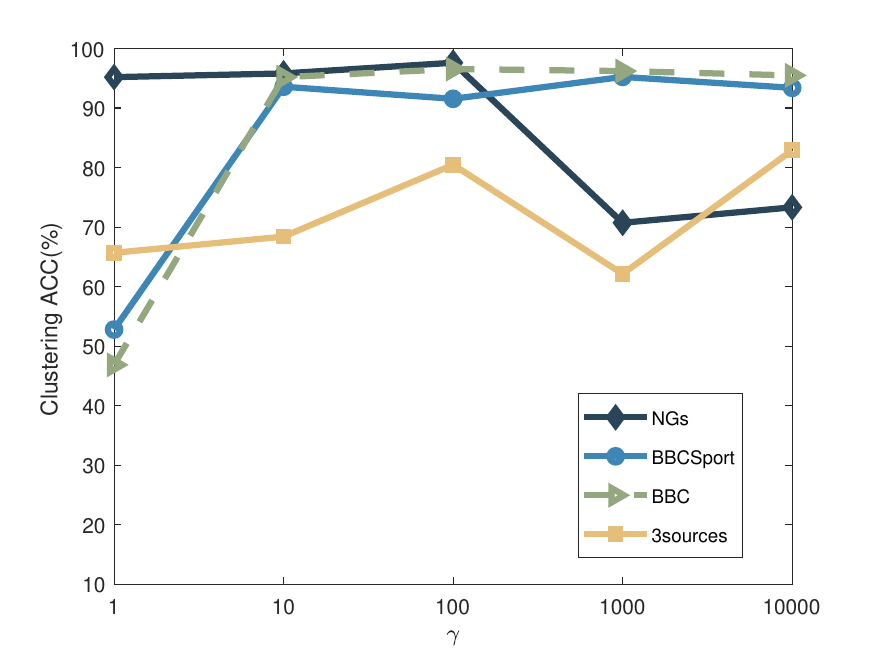}}
	\subfigure[\label{Fig:18}]{%
		\includegraphics[scale=0.32]{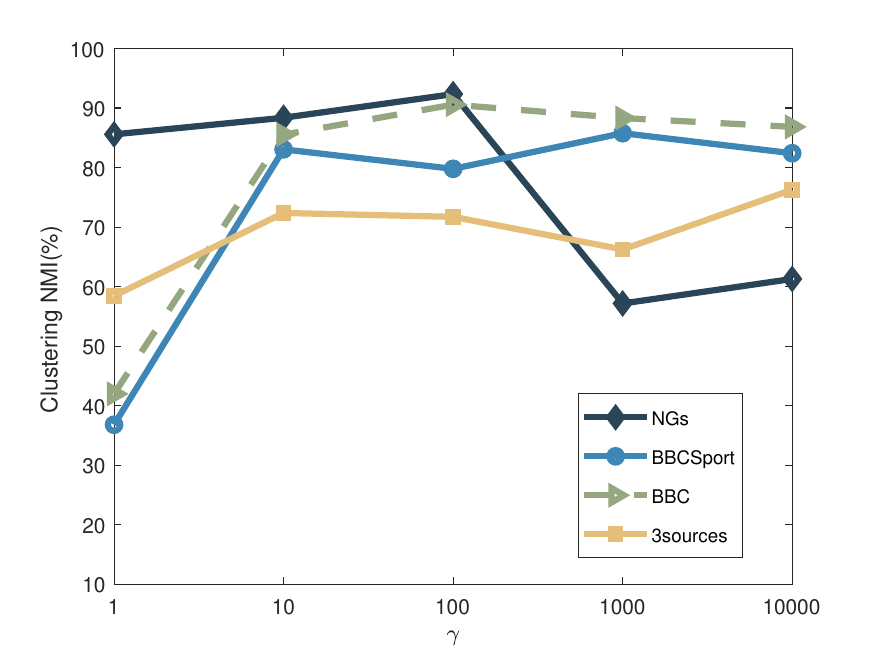}}
	\subfigure[\label{Fig:19}]{%
		\includegraphics[scale=0.32]{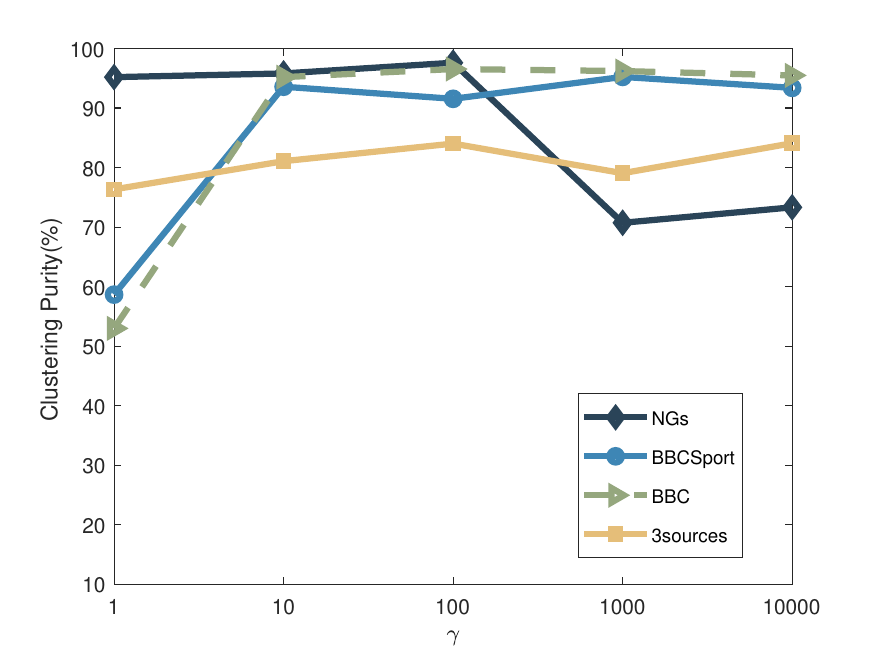}}
	
	\caption{Clustering ACC$(\%)$, NMI$(\%)$ and Purity$(\%)$ of SMVCF with different $\gamma$ on NGs, BBCSport, BBC, and 3sources datasets}
	\label{Fig:4-3}
\end{figure*}
\begin{figure*}[htbp]
	\centering	
	\subfigure[\label{Fig:23}]{%
		\includegraphics[scale=0.32]{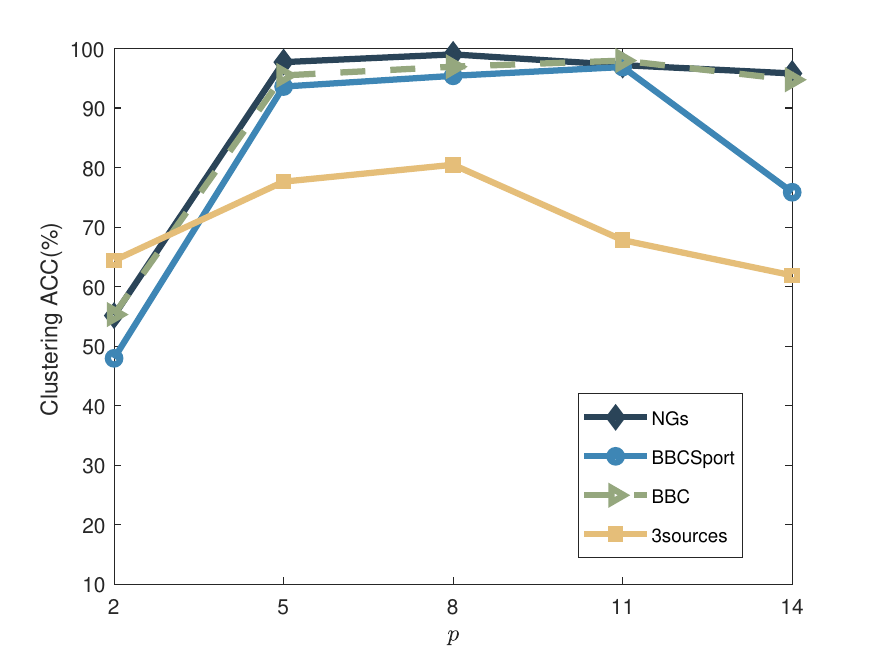}}
	\subfigure[\label{Fig:24}]{%
		\includegraphics[scale=0.32]{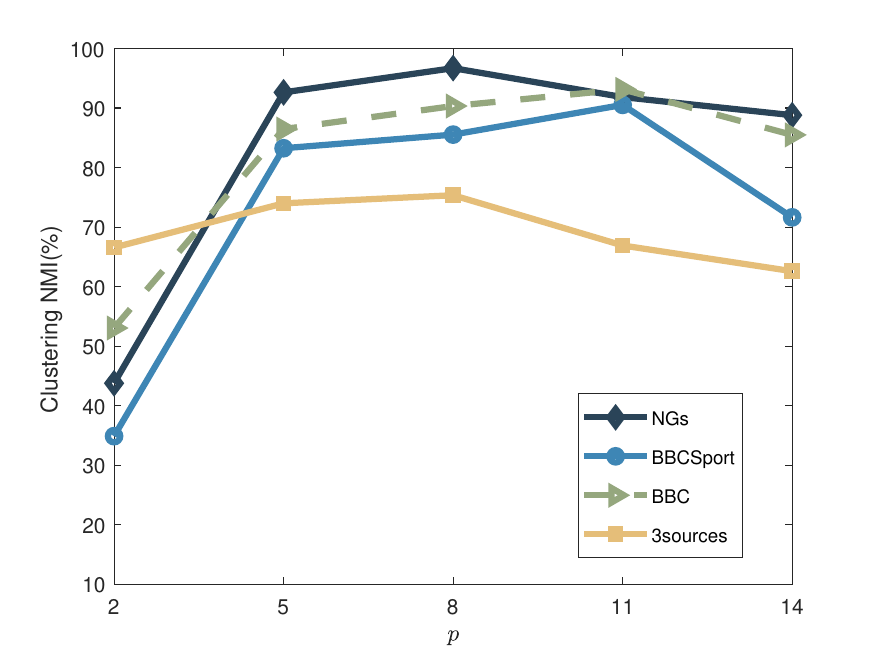}	}	
	\subfigure[\label{Fig:25}]{%
		\includegraphics[scale=0.32]{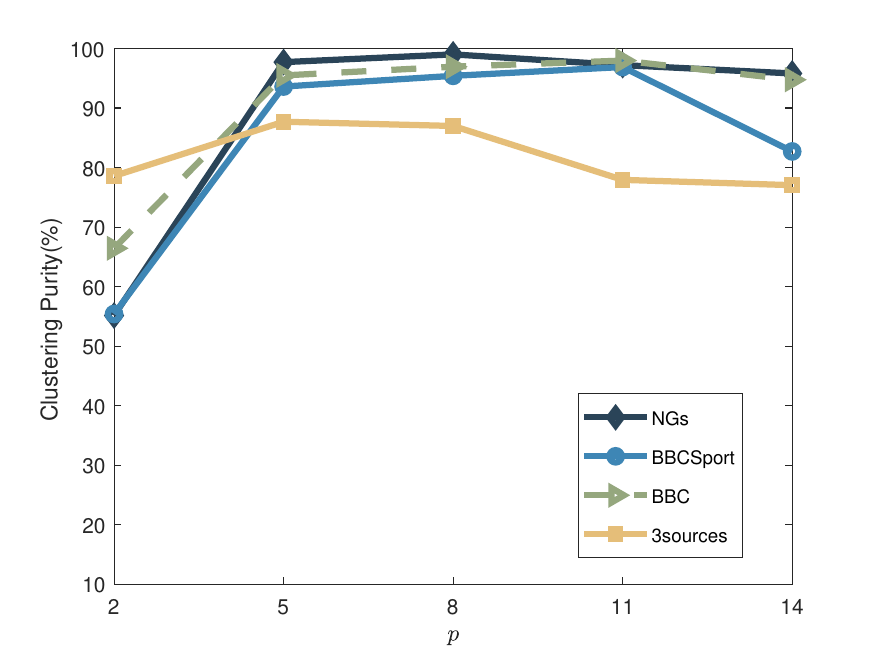}}
	
	\caption{Clustering ACC$(\%)$, NMI$(\%)$ and Purity$(\%)$ of SMVCF with different $p$ on NGs, BBCSport, BBC, and 3sources datasets}
	\label{Fig:4-4}
\end{figure*}
To test the impact of parameter variations on SMVCF, we conducted sensitivity experiments on the NGs, BBCSport, BBC, and 3sources datasets. By analyzing the SMVCF model, we identified the following parameters: (1) Explicit parameters: $\lambda$, $\beta$, and $\gamma$. Here, $\lambda$ is the coefficient for graph regularization, controlling the strength of the graph constraint. $\beta$ balances the relationship between the SMVCF reconstruction term and the label propagation term. $\gamma$ controls the weight distribution among different views. (2) Implicit parameter: the number of nearest neighbors $p$ for the undirected graph. We first analyzed the explicit parameters and then performed targeted analysis for the implicit parameter.

(1) Analysis of explicit parameters:

We divided the experiments into three scenarios:
\begin{itemize}
	\item [1)]
	Fixing $\beta$=1 and $\gamma$=100, with a label ratio set at$20\%$, we searched for the optimal parameter within the range of $\lambda \in[1,10,100,1000,10000]$.
\end{itemize}
\begin{itemize}
	\item [2)]
	Fixing $\lambda$=1 and $\gamma$=100, with a label ratio set at $20\%$, we searched for the optimal parameter within the range of $\beta \in[0.01,0.1,1,10,100]$.
\end{itemize}
\begin{itemize}
	\item [3)]
	Fixing $\lambda$=1 and $\beta$=1, with a label ratio set at $20\%$, we searched for the optimal parameter within the range of $\gamma\in[1,10,100,1000,10000]$.
\end{itemize}
For each experimental result, we ran SMVCF 10 times and reported its average performance. The specific experimental results can be seen in Figure \ref{Fig:4-1}, Figure \ref{Fig:4-2}, and Figure \ref{Fig:4-3}.

From Figure \ref{Fig:4-1}, it can be observed that when $\lambda$ varies within the experimental range, SMVCF exhibits relatively stable performance in terms of ACC, NMI, and Purity. Moreover, at $\lambda=1$, SMVCF achieves the best performance across the NGs, BBC, and 3sources datasets.

Figure \ref{Fig:4-2} shows that SMVCF maintains a high level of stability in ACC, NMI, and Purity metrics within the range of $\beta \in[0.01,0.1,1,10]$. In most cases, SMVCF performs optimally when $\beta=1$.

In Figure \ref{Fig:4-3}, it is evident that SMVCF's performance in ACC, NMI, and Purity exhibits noticeable fluctuations as $\gamma$ varies. This is expected since $\gamma$ is the parameter that influences the weight distribution among different views. Specifically, when $\gamma=100$, we generally obtain relatively excellent performance. However, both excessively large and small values of $\gamma$ have varying degrees of impact on the clustering performance, highlighting the crucial role of multi-view weight allocation in the model.

(2) Implicit Parameter Analysis:

The number of nearest neighbors in the undirected graph, denoted as $p$, is also related to label propagation, and its variation can affect the clustering performance of the SMVCF model. We tested the impact of $p$ on clustering performance on different datasets, while keeping $\lambda=1$, $\beta=1$, $\gamma=100$, and the label proportion set to $20\%$. We explored the range of $p\in[2,5,8,11,14]$ to find the optimal parameter. The detailed experimental results can be seen in Figure \ref{Fig:4-4}. It is evident that within the range of $p\in[5,8,11]$, SMVCF's semi-supervised clustering performance exhibits relatively low fluctuations in terms of ACC, NMI, and Purity. Considering stability, we suggest setting $p$ to 5.

By analyzing the above experimental results, we recommend setting $\lambda=1$, $\beta=1$, $\gamma=100$, and $p=5$ as the default values for the SMVCF model.
\subsection{Results and Analysis}
\begin{table*}[htbp]
	\caption{Average and standard deviation value of ACC$(\%)$ of different methods on NGs, BBCSport, BBC and 3sources datasets. Bolded numbers indicate the best results.}
	\label{table:42}
	\centering
	\scalebox{1.0}{
		\begin{tabular}{cc|ccccccc}
			\hline
			Datasets              & Label Ratio   & DICS            & PSLF$^{k}$          & PSLF$^{w}$          & GPSNMF$^{k}$        & GPSNMF$^{w}$        & MVSL            & SMVCF                   \\ \hline
			\multicolumn{1}{c|}{\multirow{4}{*}{NGs}} &
			10$\%$ &
			34.76$\pm$2.33 &
			27.78$\pm$2.28 &
			38.62$\pm$6.13 &
			38.22$\pm$3.26 &
			65.78$\pm$5.37 &
			70.44$\pm$4.60 &
			\textbf{95.84$\pm$0.22} \\
			\multicolumn{1}{c|}{} & 20$\%$ & 54.00$\pm$4.56  & 28.75$\pm$2.02 & 39.20$\pm$7.68 & 34.75$\pm$5.33 & 76.75$\pm$4.31 & 88.95$\pm$4.65  & \textbf{95.60$\pm$1.24} \\
			\multicolumn{1}{c|}{} & 30$\%$ & 54.91$\pm$3.64  & 32.86$\pm$4.49 & 47.89$\pm$6.01 & 28.00$\pm$6.16 & 79.14$\pm$4.00 & 89.89$\pm$3.76  & \textbf{98.40$\pm$0.00} \\
			\multicolumn{1}{c|}{} & 40$\%$ & 64.13$\pm$3.96  & 35.47$\pm$4.73 & 45.07$\pm$7.13 & 28.33$\pm$2.41 & 87.67$\pm$3.44 & 93.13$\pm$1.07  & \textbf{99.60$\pm$1.24} \\ \hline
			\multicolumn{1}{c|}{\multirow{4}{*}{BBCSport}} &
			10$\%$ &
			44.73$\pm$10.13 &
			39.18$\pm$3.59 &
			43.88$\pm$5.93 &
			43.88$\pm$4.89 &
			70.00$\pm$4.58 &
			76.23$\pm$10.47 &
			\textbf{95.40$\pm$0.00} \\
			\multicolumn{1}{c|}{} & 20$\%$ & 60.92$\pm$4.27  & 45.75$\pm$4.98 & 45.06$\pm$2.82 & 38.62$\pm$7.83 & 77.70$\pm$3.75 & 88.79 $\pm$1.30 & \textbf{96.14$\pm$1.24} \\
			\multicolumn{1}{c|}{} & 30$\%$ & 68.98$\pm$4.94  & 47.14$\pm$6.51 & 46.56$\pm$5.62 & 36.75$\pm$4.54 & 81.36$\pm$4.40 & 90.21$\pm$3.54  & \textbf{96.14$\pm$1.24} \\
			\multicolumn{1}{c|}{} & 40$\%$ & 64.13$\pm$3.96  & 46.44$\pm$2.83 & 43.99$\pm$5.54 & 34.66$\pm$4.57 & 80.06$\pm$2.65 & 92.04$\pm$2.68  & \textbf{97.43$\pm$0.00} \\ \hline
			\multicolumn{1}{c|}{\multirow{4}{*}{BBC}} &
			10$\%$ &
			52.16$\pm$11.82 &
			41.10$\pm$7.38 &
			39.58$\pm$6.52 &
			38.31$\pm$4.80 &
			62.01$\pm$4.93 &
			\textbf{94.45$\pm$0.67} &
			93.28$\pm$2.34 \\
			\multicolumn{1}{c|}{} & 20$\%$ & 74.49$\pm$5.20  & 39.01$\pm$7.43 & 39.78$\pm$3.89 & 44.71$\pm$5.92 & 77.74$\pm$3.59 & 94.55$\pm$0.56  & \textbf{95.65$\pm$0.00} \\
			\multicolumn{1}{c|}{} & 30$\%$ & 72.53$\pm$11.40 & 39.92$\pm$4.11 & 42.38$\pm$7.28 & 46.97$\pm$8.52 & 77.24$\pm$3.36 & 94.54$\pm$0.99  & \textbf{97.23$\pm$0.00} \\
			\multicolumn{1}{c|}{} & 40$\%$ & 72.12$\pm$7.76  & 37.13$\pm$8.51 & 46.13$\pm$4.04 & 55.72$\pm$5.59 & 76.16$\pm$2.85 & 83.57$\pm$9.88  & \textbf{98.54$\pm$0.00} \\ \hline
			\multicolumn{1}{c|}{\multirow{4}{*}{3sources}} &
			10$\%$ &
			38.03$\pm$1.83 &
			45.13$\pm$3.68 &
			33.42$\pm$2.81 &
			34.21$\pm$7.42 &
			34.87$\pm$12.12 &
			56.56$\pm$7.37 &
			\textbf{73.61$\pm$1.80} \\
			\multicolumn{1}{c|}{} & 20$\%$ & 45.78$\pm$5.63  & 50.81$\pm$2.75 & 37.48$\pm$6.32 & 34.07$\pm$8.17 & 70.37$\pm$5.52 & 61.05$\pm$7.08  & \textbf{81.07$\pm$0.01} \\
			\multicolumn{1}{c|}{} & 30$\%$ & 50.00$\pm$6.78  & 46.95$\pm$7.55 & 40.00$\pm$9.63 & 42.37$\pm$4.98 & 73.73$\pm$5.67 & 57.85$\pm$6.59  & \textbf{88.17$\pm$8.10} \\
			\multicolumn{1}{c|}{} & 40$\%$ & 54.65$\pm$3.24  & 45.15$\pm$3.33 & 39.80$\pm$9.03 & 42.37$\pm$4.98 & 73.73$\pm$5.67 & 56.50$\pm$12.10 & \textbf{96.45$\pm$0.00} \\ \hline
	\end{tabular}}
	\centering
\end{table*}

\begin{table*}[htbp]
	\caption{Average and standard deviation value of NMI$(\%)$ of different methods on NGs, BBCSport, BBC and 3sources datasets. Bolded numbers indicate the best results.}
	\label{table:43}
	\centering
	\scalebox{1.0}{
		\begin{tabular}{cc|ccccccc}
			\hline
			Datasets              & Label Ratio   & DICS           & PSLF$^{k}$          & PSLF$^{w}$           & GPSNMF$^{k}$        & GPSNMF$^{w}$        & MVSL           & SMVCF                   \\ \hline
			\multicolumn{1}{c|}{\multirow{4}{*}{NGs}} &
			10$\%$ &
			16.38$\pm$3.24 &
			11.70$\pm$2.15 &
			14.91$\pm$3.94 &
			23.56$\pm$3.40 &
			32.50$\pm$6.41 &
			56.07$\pm$6.85 &
			\textbf{87.50$\pm$0.69} \\
			\multicolumn{1}{c|}{} & 20$\%$ & 36.70$\pm$6.58 & 11.76$\pm$2.01 & 30.25$\pm$2.35  & 21.25$\pm$5.29 & 48.78$\pm$5.13 & 77.64$\pm$5.99 & \textbf{88.37$\pm$9.29} \\
			\multicolumn{1}{c|}{} & 30$\%$ & 43.34$\pm$3.64 & 18.12$\pm$6.02 & 21.18$\pm$6.08  & 18.59$\pm$5.30 & 53.27$\pm$5.92 & 79.72$\pm$5.52 & \textbf{95.01$\pm$0.00} \\
			\multicolumn{1}{c|}{} & 40$\%$ & 49.57$\pm$3.17 & 17.00$\pm$7.87 & 22.15$\pm$6.43  & 21.57$\pm$2.18 & 69.57$\pm$5.91 & 84.30$\pm$2.62 & \textbf{98.60$\pm$1.24} \\ \hline
			\multicolumn{1}{c|}{\multirow{4}{*}{BBCSport}} &
			10$\%$ &
			22.70$\pm$12.15 &
			14.07$\pm$2.28 &
			14.61$\pm$6.68 &
			25.85$\pm$6.25 &
			39.02$\pm$5.75 &
			57.86$\pm$8.41 &
			\textbf{86.06$\pm$0.00} \\
			\multicolumn{1}{c|}{} & 20$\%$ & 47.51$\pm$7.33 & 22.54$\pm$5.36 & 20.43$\pm$3.90  & 22.28$\pm$7.20 & 52.56$\pm$6.14 & 72.10$\pm$2.43 & \textbf{88.46$\pm$1.24} \\
			\multicolumn{1}{c|}{} & 30$\%$ & 56.51$\pm$4.83 & 24.23$\pm$5.61 & 21.31$\pm$6.45  & 17.74$\pm$5.41 & 58.57$\pm$7.59 & 74.91$\pm$5.40 & \textbf{88.35$\pm$0.00} \\
			\multicolumn{1}{c|}{} & 40$\%$ & 49.57$\pm$3.17 & 24.18$\pm$1.16 & 19.74$\pm$3.11  & 24.48$\pm$4.18 & 59.06$\pm$3.73 & 78.19$\pm$5.33 & \textbf{91.57$\pm$0.00} \\ \hline
			\multicolumn{1}{c|}{\multirow{4}{*}{BBC}} &
			10$\%$ &
			32.09$\pm$10.80 &
			16.01$\pm$10.52 &
			6.73$\pm$3.84 &
			28.35$\pm$4.20 &
			26.68$\pm$5.95 &
			\textbf{83.33$\pm$1.78} &
			81.24$\pm$0.00 \\
			\multicolumn{1}{c|}{} & 20$\%$ & 57.30$\pm$6.83 & 17.02$\pm$9.60 & 8.72$\pm$2.24   & 26.81$\pm$5.44 & 50.24$\pm$4.78 & 83.38$\pm$1.34 & \textbf{86.60$\pm$0.00} \\
			\multicolumn{1}{c|}{} & 30$\%$ & 57.15$\pm$9.19 & 13.37$\pm$4.77 & 11.11$\pm$5.50  & 27.07$\pm$7.26 & 47.28$\pm$5.46 & 82.98$\pm$2.26 & \textbf{90.47$\pm$0.00} \\
			\multicolumn{1}{c|}{} & 40$\%$ & 58.08$\pm$4.75 & 13.79$\pm$6.96 & 13.44$\pm$3.07  & 41.06$\pm$8.15 & 47.24$\pm$4.96 & 76.58$\pm$5.99 & \textbf{94.56$\pm$0.00} \\ \hline
			\multicolumn{1}{c|}{\multirow{4}{*}{3sources}} &
			10$\%$ &
			28.80$\pm$3.48 &
			34.21$\pm$7.81 &
			6.25$\pm$3.36 &
			14.36$\pm$6.84 &
			26.43$\pm$5.04 &
			52.37$\pm$5.77 &
			\textbf{67.90$\pm$1.96} \\
			\multicolumn{1}{c|}{} & 20$\%$ & 37.58$\pm$2.89 & 37.58$\pm$8.3  & 11$\pm$7.91     & 17.16$\pm$8.47 & 45.33$\pm$7.21 & 56.06$\pm$5.88 & \textbf{77.69$\pm$0.02} \\
			\multicolumn{1}{c|}{} & 30$\%$ & 46.40$\pm$9.89 & 33.16$\pm$9.78 & 12.91$\pm$11.95 & 24.64$\pm$6.87 & 49.45$\pm$7.40 & 55.90$\pm$5.25 & \textbf{86.11$\pm$5.40} \\
			\multicolumn{1}{c|}{} & 40$\%$ & 49.17$\pm$5.02 & 34.42$\pm$5.93 & 16.13$\pm$5.96  & 26.64$\pm$6.87 & 49.45$\pm$7.40 & 53.81$\pm$8.43 & \textbf{90.92$\pm$0.00} \\ \hline
	\end{tabular}}
	\centering
\end{table*}

\begin{table*}[htbp]
	\caption{Average and standard deviation value of Purity$(\%)$ of different methods on NGs, BBCSport, BBC and 3sources datasets. Bolded numbers indicate the best results.}
	\label{table:44}
	\centering
	\scalebox{1.0}{
		\begin{tabular}{cc|ccccccc}
			\hline
			Datasets &
			Label Ratio &
			DICS &
			PSLF$^{k}$ &
			PSLF$^{w}$ &
			GPSNMF$^{k}$ &
			GPSNMF$^{w}$ &
			MVSL &
			SMVCF \\ \hline
			\multicolumn{1}{c|}{\multirow{4}{*}{NGs}} &
			10$\%$ &
			35.29$\pm$2.42 &
			28.53$\pm$1.14 &
			38.62$\pm$6.13 &
			41.11$\pm$3.64 &
			65.78$\pm$5.37 &
			70.44$\pm$4.60 &
			\textbf{95.84$\pm$0.22} \\
			\multicolumn{1}{c|}{} &
			20$\%$ &
			54.25$\pm$4.26 &
			30.25$\pm$2.35 &
			40$\pm$6.99 &
			34.75$\pm$5.19 &
			76.75$\pm$4.31 &
			88.95$\pm$4.65 &
			\textbf{95.60$\pm$1.24} \\
			\multicolumn{1}{c|}{} &
			30$\%$ &
			55.03$\pm$3.75 &
			33.94$\pm$4.50 &
			47.89$\pm$6.01 &
			29.43$\pm$5.55 &
			79.14$\pm$4.00 &
			89.89$\pm$3.76 &
			\textbf{98.40$\pm$0.00} \\
			\multicolumn{1}{c|}{} &
			40$\%$ &
			64.13$\pm$3.96 &
			37.27$\pm$5.25 &
			45.87$\pm$6.98 &
			32.00$\pm$1.87 &
			87.67$\pm$3.44 &
			93.13$\pm$1.07 &
			\textbf{99.60$\pm$1.24} \\ \hline
			\multicolumn{1}{c|}{\multirow{4}{*}{BBCSport}} &
			10$\%$ &
			52.20$\pm$9.18 &
			43.67$\pm$1.76 &
			43.88$\pm$5.93 &
			47.14$\pm$3.98 &
			70.00$\pm$4.58 &
			76.84$\pm$9.51 &
			\textbf{95.40$\pm$0.00} \\
			\multicolumn{1}{c|}{} &
			20$\%$ &
			65.89$\pm$3.88 &
			47.26$\pm$4.53 &
			45.06$\pm$2.82 &
			41.84$\pm$7.40 &
			77.70$\pm$3.75 &
			88.79$\pm$1.30 &
			\textbf{96.14$\pm$1.24} \\
			\multicolumn{1}{c|}{} &
			30$\%$ &
			70.81$\pm$3.83 &
			50.24$\pm$5.07 &
			46.56$\pm$5.62 &
			41.73$\pm$3.35 &
			81.36$\pm$4.40 &
			90.21$\pm$3.54 &
			\textbf{96.14$\pm$1.24} \\
			\multicolumn{1}{c|}{} &
			40$\%$ &
			64.13$\pm$3.96 &
			47.61$\pm$2.79 &
			43.99$\pm$5.54 &
			38.96$\pm$3.84 &
			80.06$\pm$2.65 &
			90.04$\pm$2.68 &
			\textbf{97.43$\pm$0.00} \\ \hline
			\multicolumn{1}{c|}{\multirow{4}{*}{BBC}} &
			10$\%$ &
			56.56$\pm$7.98 &
			45.16$\pm$7.55 &
			39.64$\pm$6.46 &
			49.19$\pm$3.37 &
			62.01$\pm$4.81 &
			\textbf{94.45$\pm$0.67} &
			93.28$\pm$2.34 \\
			\multicolumn{1}{c|}{} &
			20$\%$ &
			74.96$\pm$4.43 &
			43.28$\pm$7.26 &
			39.78$\pm$3.89 &
			47.63$\pm$5.65 &
			77.74$\pm$3.59 &
			94.55$\pm$1.56 &
			\textbf{95.62$\pm$0.00} \\
			\multicolumn{1}{c|}{} &
			30$\%$ &
			74.99$\pm$6.33 &
			43.63$\pm$4.67 &
			42.38$\pm$7.28 &
			49.06$\pm$7.67 &
			77.24$\pm$3.36 &
			94.54$\pm$0.99 &
			\textbf{97.23$\pm$0.00} \\
			\multicolumn{1}{c|}{} &
			40$\%$ &
			76.20$\pm$3.77 &
			45.55$\pm$6.44 &
			46.13$\pm$4.04 &
			56.93$\pm$4.28 &
			76.16$\pm$2.85 &
			91.90$\pm$2.54 &
			\textbf{98.54$\pm$0.00} \\ \hline
			\multicolumn{1}{c|}{\multirow{4}{*}{3sources}} &
			10$\%$ &
			55.13$\pm$2.56 &
			55.66$\pm$5.32 &
			35.53$\pm$2.08 &
			39.47$\pm$6.66 &
			60.53$\pm$5.34 &
			68.21$\pm$6.51 &
			\textbf{77.87$\pm$1.85} \\
			\multicolumn{1}{c|}{} &
			20$\%$ &
			61.33$\pm$3.49 &
			58.96$\pm$3.13 &
			38.81$\pm$4.87 &
			40.00$\pm$8.50 &
			70.37$\pm$5.51 &
			72.93$\pm$4.05 &
			\textbf{88.05$\pm$0.01} \\
			\multicolumn{1}{c|}{} &
			30$\%$ &
			66.78$\pm$8.18 &
			57.63$\pm$7.79 &
			40.51$\pm$9.50 &
			44.92$\pm$6.90 &
			73.73$\pm$5.26 &
			72.89$\pm$4.35 &
			\textbf{93.49$\pm$3.24} \\
			\multicolumn{1}{c|}{} &
			40$\%$ &
			68.32$\pm$3.50 &
			57.23$\pm$5.49 &
			41.19$\pm$6.52 &
			44.92$\pm$6.90 &
			73.73$\pm$5.26 &
			69.13$\pm$7.05 &
			\textbf{96.45$\pm$0.00} \\ \hline
	\end{tabular}}
	\centering
\end{table*}

In this section, we compare the performance of SMVCF with four other semi-supervised multi-view clustering models (DICS, PSLF, GPSNMF and MVSL) on four publicly available multi-view datasets. It is worth noting that, according to the literature \cite{liu2014partially}, for the partially structure-sharing methods PSLF and GPSNMF, we set the dimension of the partially shared latent representation, denoted as $K$, to 100 and set the common factor ratio $\lambda=0.5$. Specifically, we have $K_{c}+K_{s} \times P=100, \quad K_{c} /\left(K_{s}+K_{c}\right)=0.5$. In the following experiments, PSLF$^{w}$ and GPSNMF$^{w}$ use the regression coefficient matrix $\mathbf{W}$ to obtain clustering labels. Given the latent factor $\mathbf{v}_i$, the clustering label $y$ is computed as $y={\arg \max} _{c} y_{c, i}$ where $y_{i}=\mathbf{W}^{\top} \mathbf{v}_{i} $. The remaining algorithms, including DICS, PSLF$^{k}$, GPSNMF$^{k}$, MVSL, and SMVCF, obtain clustering labels through $K-means$ clustering using the obtained latent representations.

Regarding the parameter settings, for SMVCF, we set $\lambda=1$, $\beta=1$, $\gamma=100$ and $p=5$. For the other four comparison algorithms, we follow the suggested parameter settings from the literature \cite{zhang2018multi,liu2014partially,liang2020semi,peng2019graph}. In terms of label usage, we conduct experiments in four different semi-supervised scenarios with label proportions of 10$\%$, 20$\%$, 20$\%$, and 40$\%$. The specific clustering experimental results can be found in Table \ref{table:42}, \ref{table:43} and Table \ref{table:44}.

Based on the aforementioned experimental results, we can draw the following conclusions:
\begin{itemize}
	\item [(1)]
	In the case of 10$\%$ labeled data, comparing the results of different algorithms on the four datasets, we observe that even under the constraint of limited labeled data, SMVCF demonstrates remarkable clustering performance in most cases compared to several semi-supervised multi-view methods based on the NMF framework. On the NGs dataset, we outperform the second-best algorithm MVSL, achieving approximately 25.4$\%$ improvement in ACC, 30.41$\%$ improvement in NMI, and 25.4$\%$ improvement in Purity. This further confirms the superiority of the CF framework in handling multi-view problems.
\end{itemize}

\begin{itemize}
	\item [(2)]
	In the scenarios with 20$\%$, 30$\%$ and 40$\%$ label ratios as shown in Table \ref{table:42}, Table \ref{table:43} and Table \ref{table:44}, our proposed SMVCF achieved the best performance in all metrics compared to the competing algorithms. Particularly, on the NGs dataset, SMVCF achieved exceptional performance with ACC, NMI, and Purity reaching 99.60$\%$, 98.60$\%$ and 99.60$\%$ respectively. This demonstrates the superior advantage of label propagation techniques when the amount of labeled information increases, compared to traditional semi-supervised label learning methods.
\end{itemize}

\begin{itemize}
	\item [(3)]
	In this experiment, our proposed SMVCF consistently demonstrated superior performance compared to state-of-the-art methods in most cases, and it exhibited excellent stability across different scenarios. This confirms the robustness and advancement of SMVCF. It further emphasizes the necessity of considering better label learning approaches within the context of multi-view data, under the premise of learning a more comprehensive low-dimensional representation using the CF framework.
\end{itemize}

\subsection{Convergence analysis}
\begin{figure*}[htbp]
	\centering	
	\subfigure[\label{Fig:20}]{%
		\includegraphics[scale=0.3]{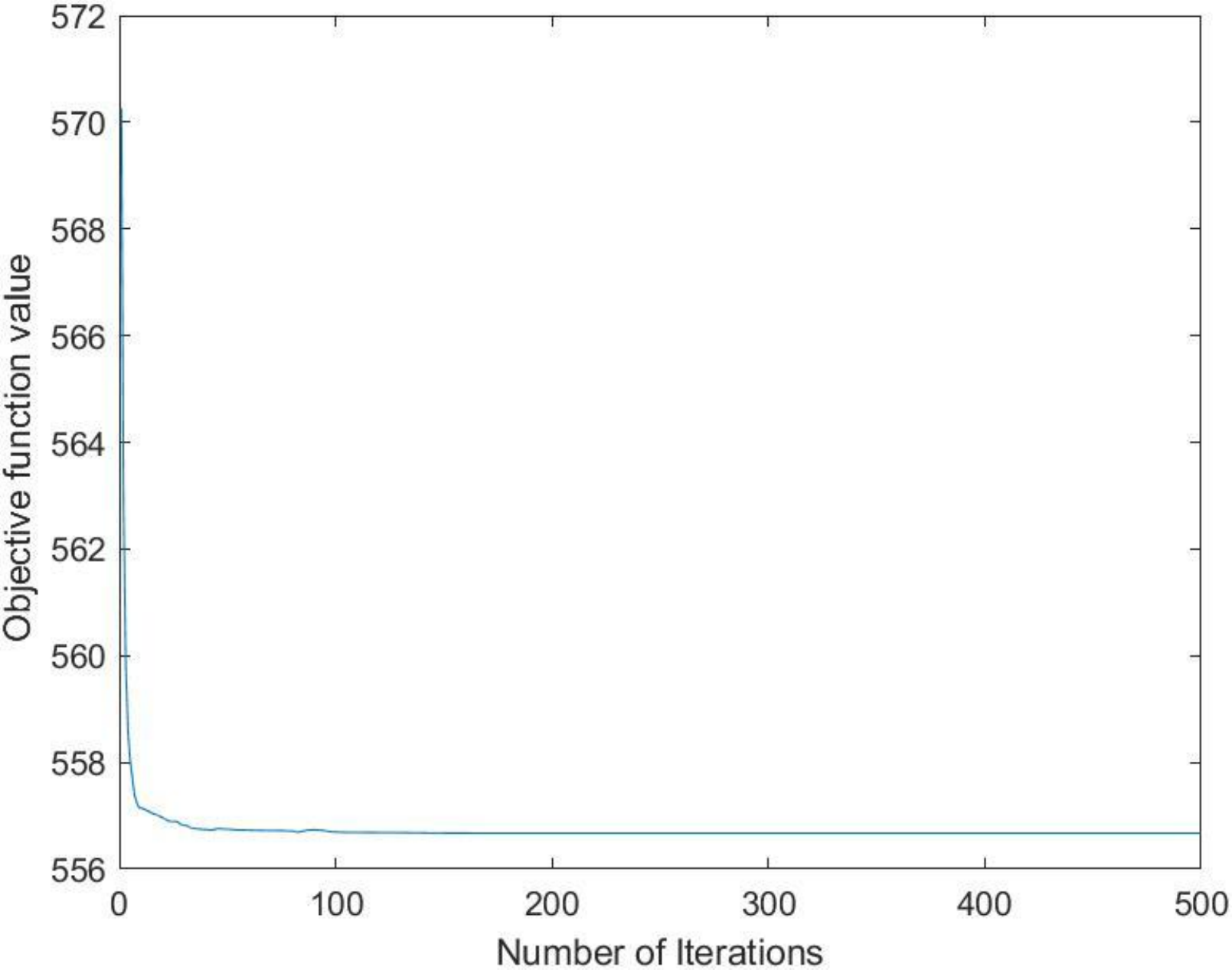}}
	\subfigure[\label{Fig:21}]{%
		\includegraphics[scale=0.3]{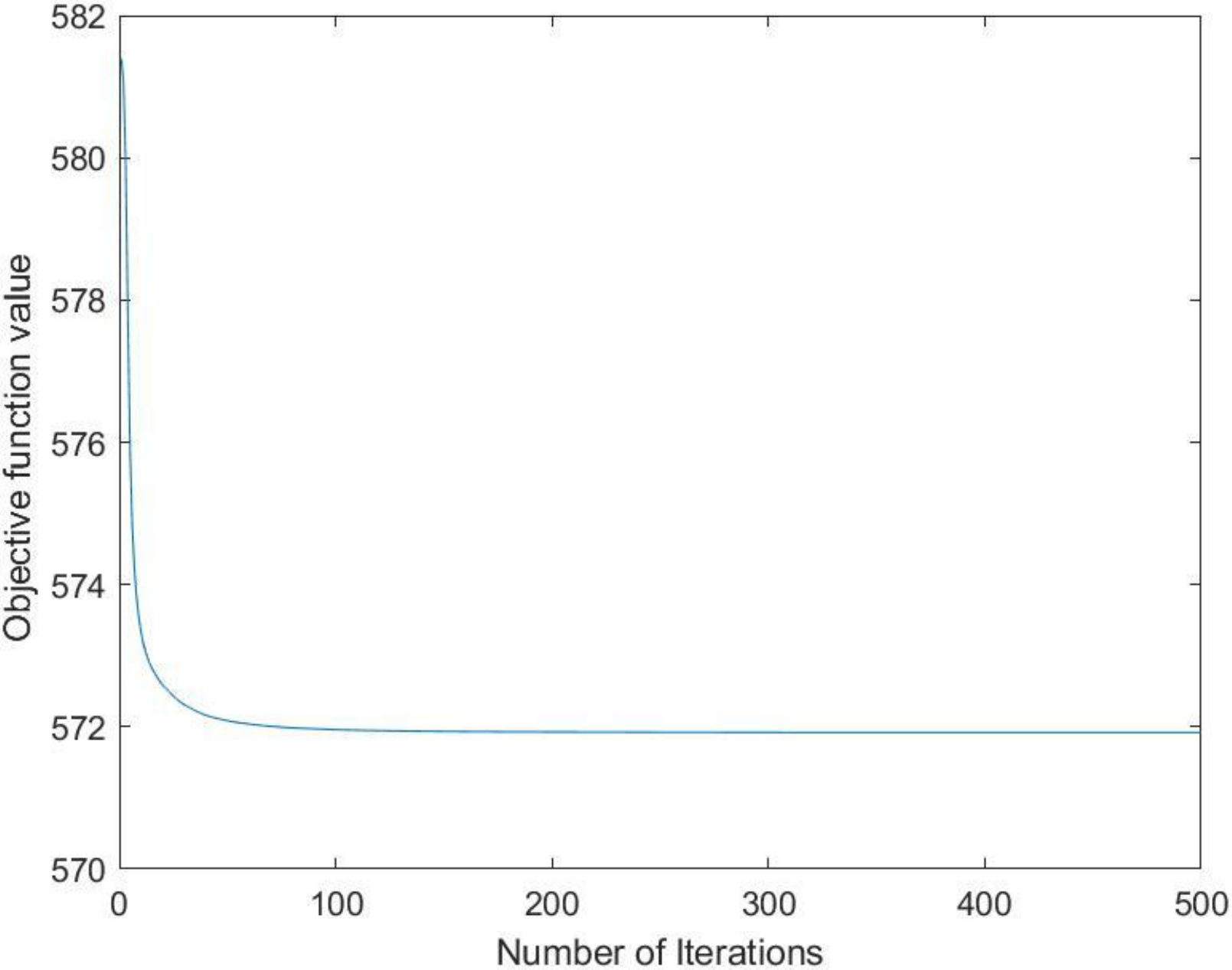}}	
	\caption{SMVCF's convergence curves on the NGs and BBCSport datasets. (a) NGs, (b) BBCSport}
	\label{Fig:4-5}	
\end{figure*}
In this section, we investigate the monotonic convergence property of SMVCF. For the experiments related to convergence, we selected NGs and BBCSport datasets as representative datasets. As shown in Figure \ref{Fig:4-5}, it can be observed that in both datasets, the SMVCF algorithm reaches convergence within approximately 30 iterations, demonstrating the excellent convergence performance of our algorithm.
\section{Conclusions}
In this paper, we propose a semi-supervised multi-view concept factorization model. Specifically, we integrate multi-view concept factorization, label propagation, and manifold learning into a unified framework to capture more useful information present in the data. Additionally, we introduce an adaptive weight vector to balance the importance of different views. Finally, we conduct extensive experiments on four different datasets with varying label proportions. The results validate the effectiveness of the SMVCF method.

\section{Acknowledgments}
This work was supported in part by the National Natural Science Foundation of China under Grant 62073087, 62071132, and 62203124.

%\bibliographystyle{cas-model2-names}
\bibliographystyle{elsarticle-num} 				
\bibliography{SMVCF-ref}

\end{document}